\documentclass[10pt,twocolumn]{article}

\usepackage{iccv}
\usepackage{times}
\usepackage{epsfig}
\usepackage{graphicx}
\usepackage{amsmath}
\usepackage{amssymb}

\usepackage{multirow}
\usepackage{footmisc}
\usepackage{color}
\usepackage[dvipsnames]{xcolor}
\usepackage{colortbl,booktabs}
\usepackage{threeparttable}
\usepackage{nopageno}
\usepackage{comment}
\usepackage{subfigure}
\usepackage{url}
\usepackage{algorithm}
\usepackage{algpseudocode}
\usepackage{mathtools}
\usepackage{pifont}
\usepackage{lipsum}
\usepackage{color,xcolor}
\definecolor{Tianlong_color}{rgb}{0.858, 0.188, 0.478}
\usepackage{adjustbox}
\usepackage{booktabs}       % professional-quality tables
\usepackage{amsfonts}       % blackboard math symbols
\usepackage{nicefrac}       % compact symbols for 1/2, etc.
\usepackage{microtype}      % microtypography
\usepackage{makecell}
\usepackage{diagbox}

\DeclarePairedDelimiterX{\inp}[2]{\langle}{\rangle}{#1, #2}

\DeclareMathAlphabet\mathbfcal{OMS}{cmsy}{b}{n}
\newcommand{\Def}[0]{\mathrel{\mathop:}=}

\definecolor{darkred}{RGB}{192, 0, 0}

\usepackage{graphicx,calc}
\newcommand*\afan{\vcenter{\hbox{\includegraphics[width=1em]{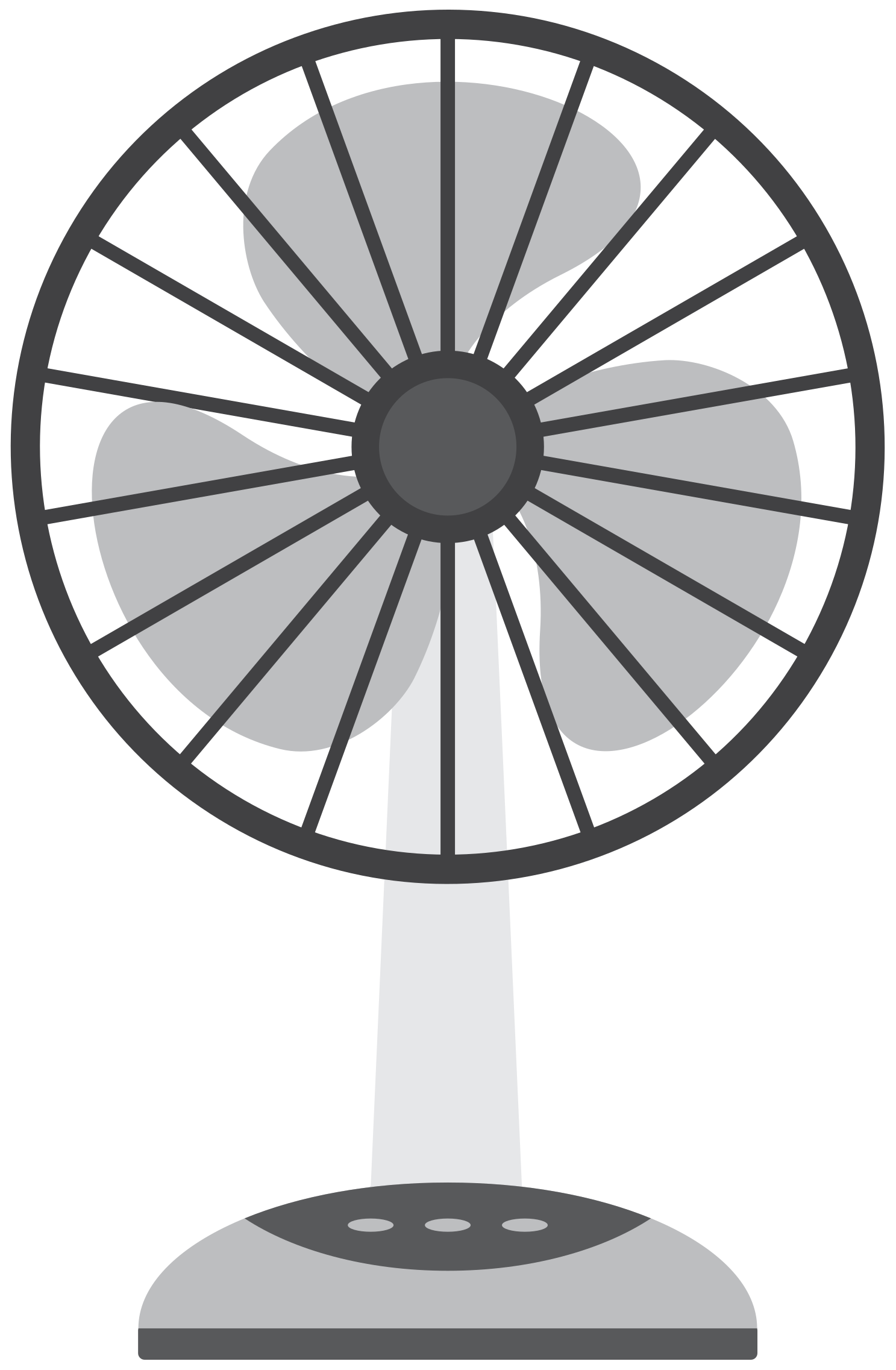}}}}

\usepackage[pagebackref=true,breaklinks=true,colorlinks,bookmarks=false]{hyperref}

\iccvfinalcopy % *** Uncomment this line for the final submission

\def\iccvPaperID{1694} % *** Enter the ICCV Paper ID here

\ificcvfinal\fi

\begin{document}

%%%%%%%%% TITLE
\title{Adversarial Feature Augmentation and Normalization for Visual Recognition}

\author{Tianlong Chen\textsuperscript{1}, Yu Cheng\textsuperscript{2}, Zhe Gan\textsuperscript{2}, Jianfeng Wang\textsuperscript{2}, Lijuan Wang\textsuperscript{2}, Zhangyang Wang\textsuperscript{1}, Jingjing Liu\textsuperscript{2}\\
\textsuperscript{1}University of Texas at Austin, \textsuperscript{2}Microsoft Corporation\\
\textit{\small \{tianlong.chen, atlaswang\}@utexas.edu,} \textit{\small \{yu.cheng, zhe.gan, jianfw, lijuanw, jingjl\}@microsoft.com}
}

\maketitle
% Remove page # from the first page of camera-ready.
\ificcvfinal\thispagestyle{empty}\fi

%%%%%%%%% ABSTRACT
\begin{abstract}
Recent advances in computer vision take advantage of adversarial data augmentation to ameliorate the generalization ability of classification models. Here, we present an effective and efficient alternative that advocates adversarial augmentation on intermediate feature embeddings, instead of relying on computationally-expensive pixel-level perturbations. We propose \textbf{A}dversarial \textbf{F}eature \textbf{A}ugmentation and \textbf{N}ormalization (A-FAN), which ($i$) first augments visual recognition models with adversarial features that integrate flexible scales of perturbation strengths, ($ii$) then extracts adversarial feature statistics from batch normalization, and re-injects them into clean features through feature normalization. We validate the proposed approach across diverse visual recognition tasks with representative backbone networks, including ResNets and EfficientNets for classification, Faster-RCNN for detection, and Deeplab V3+ for segmentation. Extensive experiments show that A-FAN yields consistent generalization improvement over strong baselines across various datasets for classification, detection and segmentation tasks, such as CIFAR-10, CIFAR-100, ImageNet, Pascal VOC2007, Pascal VOC2012, COCO2017, and Cityspaces. Comprehensive ablation studies and detailed analyses also demonstrate that adding perturbations to specific modules and layers of classification/detection/segmentation backbones yields optimal performance. Codes and pre-trained models will be made available at: \small{\url{https://github.com/VITA-Group/CV_A-FAN}}.
\end{abstract}

%%%%%%%%% BODY TEXT
\section{Introduction}

\begin{figure}[t] 
\centering
\includegraphics[width=1\linewidth]{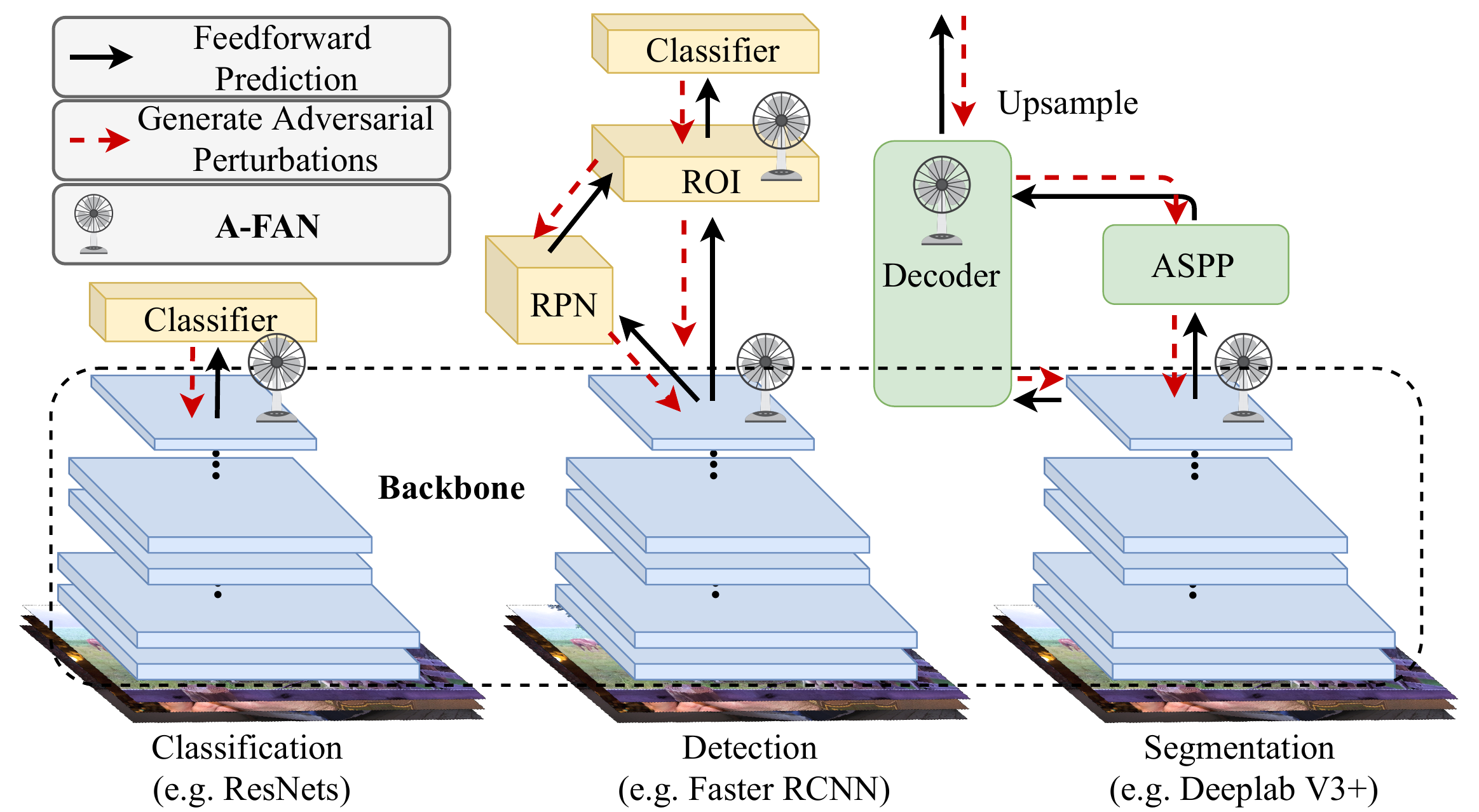}
\caption{Overview of adversarial feature augmentation and normalization ($\afan$ A-FAN) for enhanced image classification (\textit{left}), object detection (\textit{center}) and semantic segmentation (\textit{right}). We take ResNets \cite{he2016deep}, Faster RCNN \cite{ren2016faster}, and DeepLab V3+ \cite{chen2018encoder} pipelines as examples. Our proposed A-FAN mechanisms are plugged into backbone networks and/or ROI/decoder modules for classification/detection/segmentation, respectively.}
\label{fig:framework}
\vspace{-3mm}
\end{figure}

Adversarial vulnerability is a critical issue in the practical application of neural networks. Various attacks have been proposed to challenge visual recognition models of classification, detection and segmentation \cite{szegedy2013intriguing,goodfellow2014explaining,li2018robust,li2018exploring,lu2017adversarial,liu2018dpatch,lu2017adversarial,xie2017adversarial,wei2018transferable,zhang2020contextual, arnab2018robustness,shen2019advspade}. Such susceptibility has motivated abundant studies on adversarial defense mechanisms for training robust neural networks \cite{schmidt2018adversarially,sun2019towards,nakkiran2019adversarial,stutz2019disentangling,raghunathan2019adversarial,hu2019triple,chen2020adversarial,chen2021robust,jiang2020robust}, among which \textit{adversarial training} based methods \cite{madry2017towards, zhang2019theoretically}, leveraging augmented adversarial examples, have achieved consistently superior robustness than others. However, crafting high-quality adversarial examples is computationally costly, and such adversarial training often results in a negative impact on performance over clean data \cite{zhang2019theoretically}.

Interestingly, a few advanced studies turn to investigate the possibility of ameliorating networks' generalization ability via adversarial training. Recent progress shows that using adversarial perturbations to augment input data/embedding can effectively alleviate overfitting issues and lead to better generalization in multiple domains, including image classification \cite{xie2020adversarial}, language understanding \cite{wang2019improving,zhu2019freelb}, and vision-language modeling \cite{gan2020large}. However, it still suffers from expensive computational cost due to the generation of pixel-level perturbations when applied to image classification. We raise the following natural, yet largely open questions: 
% \begin{center}

% \vspace{-0.5em}
\textbf{Q1:} \textit{Can adversarial training, as data augmentation, broadly boost the performance of various visual recognition tasks on clean data, not only image classification, but also object detection, semantic segmentation or so?}

\textbf{Q2:} \textit{If the above answer is yes, can we have more efficient and effective options for adversarial data augmentation, e.g., avoiding the high cost of finding input-level adversarial perturbations?}

%we find a more efficient and effective alternative of utilizing adversarial training to improve clean accuracy of diverse visual recognition models?
% \vspace{-0.5em}
% \end{center}
%Meanwhile, such alternative approaches need to be broadly applicable, not only for classification, but also for complicated visual recognition tasks such as detection and segmentation which to our best knowledge, has never been explored.  

In this paper, we propose A-FAN (\textbf{A}dversarial \textbf{F}eature \textbf{A}ugmentation and \textbf{N}ormalization), a novel algorithm to improve the generalization for visual recognition models. Our method perturbs the representations of intermediate feature space for both task-specific modules (e.g., Classifiers for ResNets, ROI for Faster RCNN, and Decoder for Deeplab V3+) and generic backbones, as shown in Figure~\ref{fig:framework}. Specifically, A-FAN generates adversarial feature perturbations efficiently by one-step projected gradient descent, and fastly computes adversarial features with other perturbation strengths from weak to strong via interpolation. This strength-spectrum coverage allows models to consider a wide range of attack strengths simultaneously, to fully unleash the power of implicit regularization of adversarial features. 

Furthermore, A-FAN normalizes adversarial augmented features in a ``Mixup" fashion. Unlike previous work \cite{zhang2017mixup,li2020feature} that fuses inputs or features from different samples, we amalgamate adversarial and clean features by injecting adversarial statistics extracted from batch normalization into clean features. Such re-normalized features serve as an implicit label-preserving data augmentation, which smooths the learned decision surface \cite{li2020feature}. 
%Consequently, A-FAN-augmented classification, detection, and segmentation models achieve improved generalization. 
Our main contributions are summarized as follows:
\begin{itemize}
    \item We introduce a new adversarial feature augmentation approach to enhancing the generalization ability of image classification, object detection, and semantic segmentation models, by incorporating scaled perturbation strength from weak to strong simultaneously.
    \vspace{-1mm}
    \item We also propose a new feature normalization method, which extracts the statistics from adversarial perturbed features and re-injects them into the original clean features. It can be regarded as implicit label-preserving data augmentation that smooths the learned decision boundary (illustrated in Figure~\ref{fig:loss} later on). 
    \vspace{-1mm}
    \item We conduct comprehensive experiments to verify the effectiveness of our proposed approach over diverse tasks (CIFAR-10, CIFAR-100, ImageNet for image classification; Pascal VOC2007 and COCO2017 for object detection; Pascal VOC2007, Pascal VOC2012 and Cityspaces for semantic segmentation). The substantial and consistent performance lift demonstrates the superiority of our A-FAN framework.
\end{itemize}

\section{Related Work}
%\subsection{Adversarial Training}
\paragraph{Adversarial Attacks and Defenses.} When presented with adversarial samples, which are maliciously designed by imperceptible perturbations \cite{goodfellow2014explaining, kurakin2016adversarial, madry2017towards}, deep neural networks often suffer from severe performance deterioration, e.g., \cite{szegedy2013intriguing,goodfellow2014explaining,carlini2017towards,croce2020reliable} for classification models and \cite{li2018robust,li2018exploring,lu2017adversarial,liu2018dpatch,xie2017adversarial,wei2018transferable,zhang2020contextual, arnab2018robustness,shen2019advspade} for detection/segmentation models. To address this notorious vulnerability, numerous defense mechanisms \cite{zhang2019theoretically,schmidt2018adversarially,sun2019towards,nakkiran2019adversarial,stutz2019disentangling,raghunathan2019adversarial} have been proposed, such as input transformation \cite{xu2017feature, liao2018defense, guo2017countering, dziugaite2016study}, randomization \cite{liu2018adv, liu2018towards, dhillon2018stochastic}, and certified defense approaches \cite{cohen2019certified,raghunathan2018semidefinite}. Among these, adversarial-training-based methods show superior robustness in defending state-of-the-art adversarial attacks \cite{goodfellow2014explaining, kurakin2016adversarial, madry2017towards}. Although adversarial training substantially enhances model robustness, it usually comes at the price of compromising the standard accuracy~\cite{tsipras2018robustness}, which has been demonstrated both empirically and theoretically  \cite{zhang2019theoretically,schmidt2018adversarially,sun2019towards,nakkiran2019adversarial,stutz2019disentangling,raghunathan2019adversarial}.

\vspace{-3mm}
\paragraph{Adversarial Training Ameliorates Generalization.} It is unexpected, but reasonable that recent works \cite{xie2020adversarial,zhu2019freelb,wang2019improving,gan2020large,wei2019improved} present an opposite perspective: \textit{adversarial training can be leveraged to enhance models' generalization if harnessed in the right manner.} For example, \cite{xie2020adversarial} shows that image classification performance on the clean dataset can be improved by using adversarial samples with pixel-level perturbation generation. \cite{zhu2019freelb} and \cite{wang2019improving} apply adversarial training to natural language understanding and language modeling, both successfully achieving better standard accuracy. \cite{gan2020large} achieves similar success on various vision-and-language tasks. Parallel studies \cite{wei2019improved,8852250} employ handcrafted or auto-generated perturbed features to ameliorate generalization. However, adversarial training in latent feature space as a more efficient and effective alternative has, to our best knowledge, not been studied in depth, even for classification tasks. Our work comprehensively explores this possibility not only for image classification, but also for object detection and semantic segmentation which are more challenging prediction tasks and usually require a much more sophisticated model structure, posing obstacles to easily exploit adversarial information for enhanced generalization.

\begin{figure*}[t] 
\centering
\includegraphics[width=0.92\linewidth]{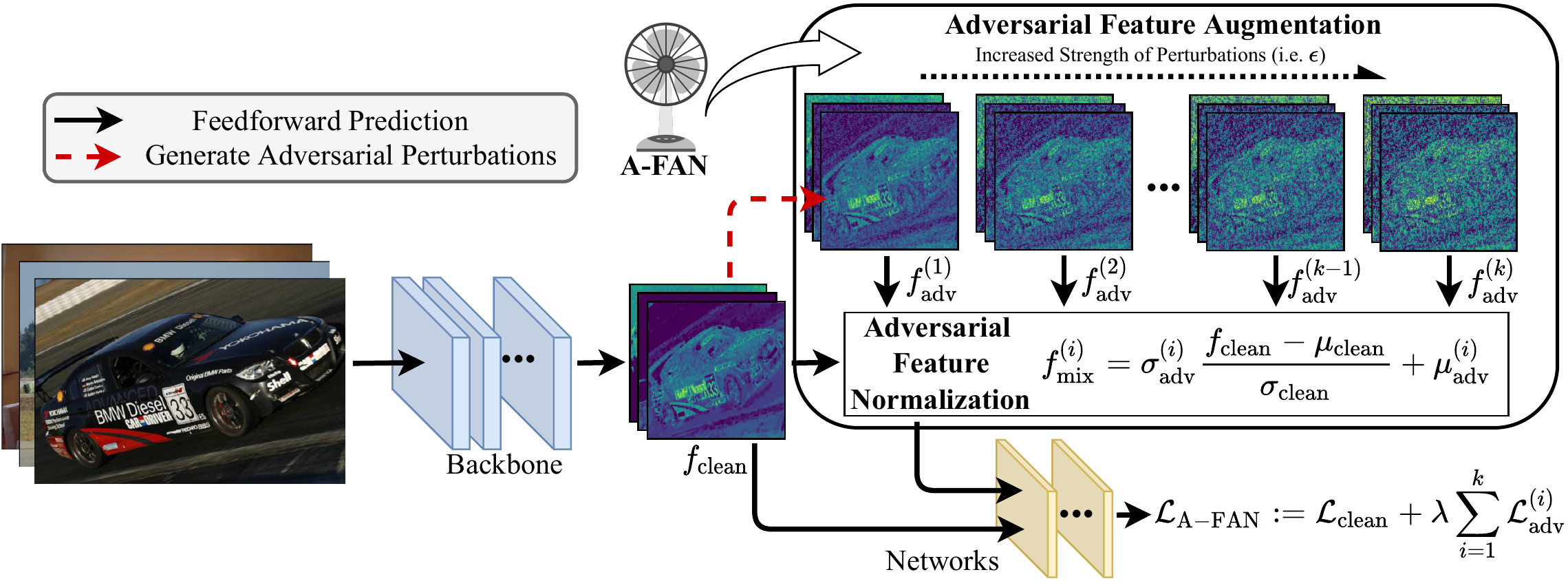}
\caption{The pipeline of $\afan$ A-FAN, which contains adversarial feature augmentation and adversarial feature normalization. From top to bottom, a series of adversarial feature perturbations with different strengths are generated to augment the intermediate clean features. Then, the statistics (i.e., $\mu_{\mathrm{adv}}$ and $\sigma_{\mathrm{adv}}$) of perturbed features $f_{\mathrm{adv}}$ are extracted and re-injected into the original clean features $f_{\mathrm{clean}}$. In the end, the normalized features $f_{\mathrm{mix}}$ are taken as inputs by the rest of the network, and optimized by $\mathcal{L}_{\mathrm{A-FAN}}$ with standard ($\mathcal{L}_{\mathrm{clean}}$) and adversarial ($\mathcal{L}_{\mathrm{adv}}$) training objectives.}
\label{fig:a-fan}
\vspace{-3mm}
\end{figure*}

\vspace{-3mm}
\paragraph{Feature Augmentation and Normalization.}
Pixel-level data augmentation techniques have been widely adopted in visual recognition models,e.g., \cite{simard1993efficient,scholkopf1996incorporating,cubuk2018autoaugment,hendrycks2019augmix} for classification, \cite{Detectron2018,liu2016ssd,zoph2019learning} for detection and segmentation. They are generic pipelines for augmenting training data with image-level information. Adversarial samples can also serve as a data augmentation method \cite{xie2020adversarial}. However, feature space augmentations have not received the same level of attention. A few pioneering works propose generative-based feature augmentation approaches for domain adaptation \cite{8578674}, imbalanced classification \cite{zhang2019feature}, and few-shot learning \cite{chen2019multi}.

Feature normalization plays an important role in neural network training \cite{ioffe2015batch,li2020feature,montavon2012neural,li1998sphering}. \cite {ioffe2015batch} proposes batch normalization to remove biases in the dataset, which can substantially improve model generalization ability. \cite{xie2020adversarial} utilizes dual batch normalization to calculate statistics of adversarial and clean samples separately, therefore obtaining promising standard accuracy. Recent investigations \cite{ba2016layer,ulyanov2016instance,wu2018group,li2019positional,li2020feature} devote a particular attention to normalizing features of each training instance individually. As an illustration, \cite{li2020feature} leverages the first and second-order moments of extracted features and re-injects these moments into features from another instance by feature normalization. Different from them, we propose to utilize feature normalization techniques to combine adversarial and clean features to smooth the learned decision surface and improve model generalization ability. 

\section{Preliminaries}

\subsection{Rationale of A-FAN}
\paragraph{Theoretical Insights.} For linear classifiers, a large output margin, the gap between predictions on the true label and the next most confident label, implies good generalization \cite{bartlett2002rademacher,koltchinskii2002empirical,hofmann2008kernel,kakade2008complexity}. Although this relationship is less clear for non-linear deep neural networks, \cite{wei2019improved} establishes a similar generalization bound associated with the ``all-layer margin" which depends on Jacobian and intermediate layer norms. Furthermore, \cite{wei2019improved} derives theoretical analyses that appropriately injecting perturbations to intermediate features encourages a large layer margin and leads to improved generalization. A parallel study \cite{wang2019improving} presents theoretical intuitions from a new perspective that introducing adversarial noises encourages the diversity of the embedding vectors, mitigates overfitting, and improves generalization for neural language models. These observations make the main cornerstone for our A-FAN approach valid. 

\begin{figure}[t!] 
\centering
%\vspace{-2.5mm}
\includegraphics[width=0.93\linewidth]{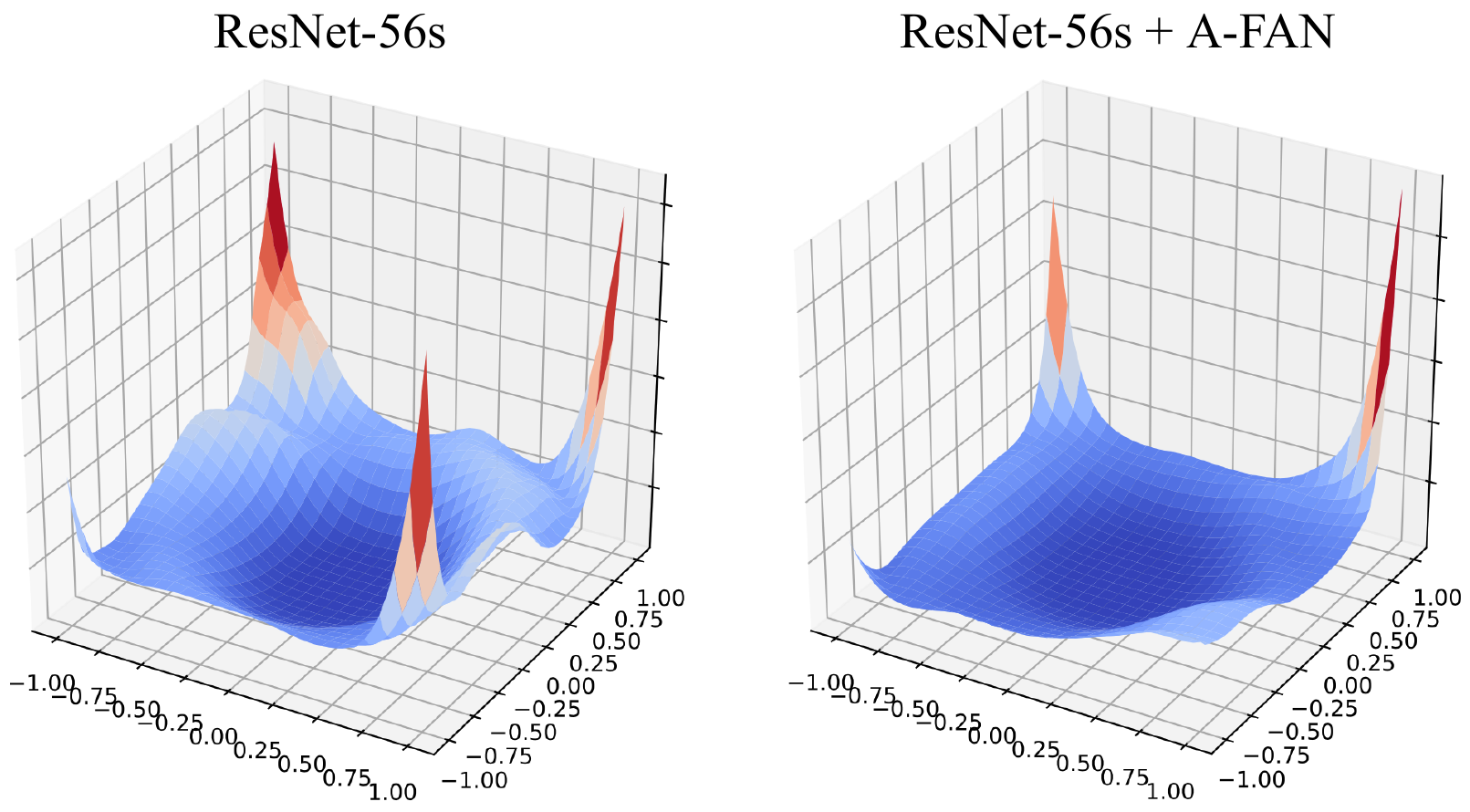}
\caption{Loss landscape of ResNet-56s with or without A-FAN on CIFAR-10. Visualization tools are provided by \cite{visualloss}.}
\label{fig:loss}
\vspace{-3mm}
\end{figure}

\vspace{-3mm}
\paragraph{Empirical Evidences.} There exist advanced studies \cite{xie2020adversarial,zhu2019freelb,wang2019improving,gan2020large,wei2019improved} revealing that appropriately utilizing adversarial perturbations ameliorates generalization ability of deep neural networks on diverse applications. Note that these designed approaches are \textit{not} defense mechanisms for adversarial robustness; instead, they serve as a special data augmentation for improved performance on clean samples. Different from input perturbations \cite{xie2020adversarial}, our work leverages adversarial perturbation in latent feature space. To further unleash the power of adversarial augmented features, we asymmetrically fuse them with clean features, which allows the model to capture and smooth out different directions of the decision boundary \cite{li2020feature}. Accordingly, A-FAN-augmented models obtain flatter loss landscape (i.e., smaller norms of Hessian with respect to model weights) and improved generalization ability, as supported in Table~\ref{table:hessian} and Figure~\ref{fig:loss}. 

\begin{table}[t!]
\centering
\caption{Performance and Hessian properties of ResNet-56s with or without A-FAN on CIFAR-10. A smaller spectral norm or trace of Hessian indicates a flatter loss landscape w.r.t. model weights.}
\label{table:hessian}
\begin{adjustbox}{width=0.47\textwidth}
\begin{tabular}{l|c|c}
\toprule
\multirow{1}{*}{Settings} & \multicolumn{1}{c}{ResNet-56s} & \multicolumn{1}{|c}{ResNet-56s + A-FAN}\\ \midrule
Standard Accuracy & 93.59 & 94.82 \\ \midrule
Spectral Norm of Hessian & 23.34 & 12.66 \\
Trace of Hessian & 246.24 & 211.94 \\
\bottomrule
\end{tabular}
\end{adjustbox}
\vspace{-3mm}
\end{table}

\subsection{Notations}
Our proposed A-FAN framework includes two key components: ($i$) adversarial feature augmentation; and ($ii$) adversarial feature normalization, as shown in Figure~\ref{fig:a-fan}. Note that we introduce adversarial perturbations in the intermediate feature space, instead of manipulating raw image pixels as in common practice. 

Let $\mathcal{D}=\{\mathbf x, \boldsymbol{y}\}$ denotes the dataset, where $\mathbf x$ is the input image and $\boldsymbol{y}$ is the corresponding ground-truth (e.g., one-hot classification labels, bounding boxes or segmentation maps). Let $f(\mathbf x;\boldsymbol{\theta})$ with $\boldsymbol{\theta}=(\boldsymbol{\theta}_{b},\boldsymbol{\theta}_{t})$ represent the predictions of neural networks, where $\boldsymbol{\theta}_b$ and $\boldsymbol{\theta}_{t}$ are the parameters of the backbone networks and task-specific modules, respectively. For example, $\boldsymbol{\theta}_t$ denotes the parameters of ResNets' classifiers; or the parameters of RPN, ROI, and classifier in Faster RCNN; or the parameters of ASPP and Decoder in Deeplab V3+. Adversarial training \cite{madry2017towards} can be formulated as follows:
\begin{align}
    \begin{array}{ll}
    \displaystyle \min_{\boldsymbol{\theta}}\mathbb E_{(\mathbf x, \boldsymbol{y}) \in \mathcal{D}} \displaystyle
\left[\max_{\|\boldsymbol{\delta}\|_{\mathrm{p}}\le\epsilon}\mathcal{L}_{\mathrm{adv}}(f(\mathbf x+\boldsymbol{\delta},\boldsymbol{\theta});\boldsymbol{y})\right],
    \end{array} \label{eq:adv}
\end{align}
where $\boldsymbol{\delta}$ is the crafted adversarial perturbation constrained within a $\ell_{\mathrm{p}}$ norm ball centered at $\mathbf x$ with a radius $\epsilon$. The radius $\epsilon$ is the maximum magnitude of generated adversarial perturbations, which roughly indicates the strength of perturbations \cite{madry2017towards}. $\mathbb E_{(\mathbf x, \boldsymbol{y}) \in \mathcal{D}}$ takes the expectation over the empirical objective over the dataset $\mathcal{D}$. The perturbation $\boldsymbol{\delta}$ can be reliably created by multi-step projected gradient descent (PGD) \cite{madry2017towards} (taking $\|\cdot\|_{\infty}$ perturbation for example):
\begin{align}\label{eq:attack_on_x}
    \begin{array}{l}
        \displaystyle\boldsymbol{\delta}_{t+1}= \Pi_{\|\boldsymbol{\delta}\|_{\infty}\le\epsilon} \displaystyle \left[ \boldsymbol{\delta}_{t}+\alpha\cdot \mathrm{sgn}(\nabla_{\mathbf x}\mathcal{L}_{\mathrm{adv}}(f(\mathbf x+\boldsymbol{\delta}_t,\boldsymbol{\theta});\boldsymbol{y}) \right],
     \end{array}
\end{align}
where $\alpha$ is the step size of inner maximization, $\mathrm{sgn}$ is the sign function, and $\mathcal{L}_{\mathrm{adv}}$ is the adversarial training objective calculated over perturbed images. 
% $t$ is the number of PGD steps, 

\subsection{Adversarial Feature Augmentation} \label{sec:afa} In this section, we present the proposed adversarial feature augmentation mechanism. Specifically, perturbations are generated in the intermediate feature space via PGD (taking features from backbone $\boldsymbol{\theta}_b$ for example):
{\small\begin{align}
    \begin{array}{ll}
    \displaystyle \min_{\boldsymbol{\theta}}\mathbb E_{(\mathbf x, \boldsymbol{y}) \in \mathcal{D}} \displaystyle
\left[\mathcal{L}_{\mathrm{clean}}+\lambda\max_{||\boldsymbol{\delta}||_{\infty}\le\epsilon}\mathcal{L}_{\mathrm{adv}}(f(\mathbf x,\boldsymbol{\theta}_b)+\boldsymbol{\delta};\boldsymbol{\theta}_t;\boldsymbol{y})\right],
    \end{array} \label{eq:afa}
\end{align}}%
where the type of $\mathcal{L}_{\mathrm{clean}}=\mathcal{L}(f(\mathbf x, \boldsymbol{\theta});\boldsymbol{y})$ and $\mathcal{L}_{\mathrm{adv}}$ are determined by tasks (e.g., detection models adopt regression and classification loss). $\lambda$ is a hyperparameter to control the influence of adversarial feature augmentation. Perturbations $\boldsymbol{\delta}$ are generated by PGD, as shown in Equation~\ref{eq:attack_on_x}, but on the features $f(\mathbf x,\boldsymbol{\theta}_b)$ from the backbone network ($\boldsymbol{\theta}_b$) rather than on raw input images. Note that the formulation in Equation~\ref{eq:afa} only considers single perturbation strength.

To fully unleash the powerful of adversarial augmentation in the feature space, we propose an enhanced technique that utilizes a series of adversarially perturbed features with strength from weak to strong simultaneously. In particular, we integrate the adversarial training objective with respect to the feature perturbation strength $\epsilon$ on an interval instead of a single point, depicted as follows:
\begin{align}
    \begin{array}{ll}
    \displaystyle \mathcal{L} = \int_0^{\mathcal{E}} \max_{||\boldsymbol{\delta}||_{\infty}\le\epsilon}\mathcal{L}_{\mathrm{adv}}(f(\mathbf x,\boldsymbol{\theta}_b)+\boldsymbol{\delta}(\epsilon);\boldsymbol{\theta}_t;\boldsymbol{y})\mathrm{d}\epsilon,
    \end{array} \label{eq:afa_interval}
\end{align}
where $[0, \mathcal{E}]$ is the integral interval for perturbation strength $\epsilon$, and $\boldsymbol{\delta}(\epsilon)$ is the crafted feature perturbation dependent on $\epsilon$. In a similar way, we can generate adversarial augmented features for the task-specific modules $\boldsymbol{\theta}_t$ in classification, detection and segmentation models.

\vspace{-3mm}
\paragraph{Approximation.} Unfortunately, the integral in Equation~\ref{eq:afa_interval} is intractable due to the lack of an explicit functional representation for deep neural networks. We provides an approximate solution by uniformly sampling $\{\epsilon^{(1)},\cdots,\epsilon^{(k)}\}\in[0,\mathcal{E}]$ and subsequently generating augmented features $\{f_{\mathrm{adv}}^{(1)},\cdots,f_{\mathrm{adv}}^{(k)}\}$, as shown in Figure~\ref{fig:a-fan}. Specifically,
\begin{align}
    %\begin{array}{ll}
    %\displaystyle   
    & \int_0^{\mathcal{E}} 
    \displaystyle \max_{||\boldsymbol{\delta}||_{\infty}\le\epsilon} \mathcal{L}_{\mathrm{adv}}(f(\mathbf x,\boldsymbol{\theta}_b)+\boldsymbol{\delta}(\epsilon);\boldsymbol{\theta}_t;\boldsymbol{y})\mathrm{d}\epsilon \label{eq:afa_app} \\ 
    \approx & \sum_{i=1}^k   %\displaystyle
    \max_{||\boldsymbol{\delta}||_{\infty}\le\epsilon^{(i)}}\mathcal{L}_{\mathrm{adv}}(f(\mathbf x,\boldsymbol{\theta}_b)+\boldsymbol{\delta}_i(\epsilon^{(i)});\boldsymbol{\theta}_t;\boldsymbol{y}), \nonumber 
    %\end{array} 
\end{align}
where $f_{\mathrm{adv}}^{(i)}=f(\mathbf x,\boldsymbol{\theta}_b)+\boldsymbol{\delta}_i(\epsilon^{(i)})$ is the adversarial augmented feature embedding. 

\subsection{Adversarial Feature Normalization}
In this section, we introduce the proposed adversarial feature normalization. Inspired by \cite{zhang2017mixup,yun2019cutmix,li2020feature}, we fuse clean ($f_{\mathrm{clean}}$) and adversarially ($f_{\mathrm{adv}}$) perturbed features for each training sample. Specifically, normalized features $f_{\mathrm{mix}}$ are crafted by normalizing clean features with adversarial feature moments. This asymmetric composition across clean and adversarial features assists networks to smooth out decision boundaries and obtain improved generalization \cite{li2020feature}. 

Let $\mu_{\mathrm{clean}}$ and $\mu_{\mathrm{adv}}^{(i)}$ denote the first-order moment of clean feature and the $i$-th augmented adversarial feature. Similarly, $\sigma_{\mathrm{clean}}$ and $\sigma_{\mathrm{adv}}^{(i)}$ denote the corresponding second-order moment. Their feature statistics are calculated in the routine of batch normalization \cite{ioffe2015batch}. Note that the statistics can also derive from other normalization approaches \cite{ba2016layer,ulyanov2016instance,wu2018group,li2019positional}, such as instance-norm. The detailed formulation is defined as follows:
\begin{align}\label{eq:afn}
    \begin{array}{l}
        \displaystyle f_{\mathrm{mix}}^{(i)}\Def \sigma_{\mathrm{adv}}^{(i)}\frac{f_{\mathrm{clean}}-\mu_{\mathrm{clean}}}{\sigma_{\mathrm{clean}}}+\mu_{\mathrm{adv}}^{(i)},
     \end{array}
\end{align}
where $i \in \{1,2,\cdots,k\}$ and $k$ are the number of augmented features. Normalized features $f_{\mathrm{mix}}^{(i)}$ are fed to the networks and computed as the adversarial training objective $\mathcal{L}_{\mathrm{adv}}^{(i)}$.

\subsection{Overall Framework of A-FAN}
As presented in Figure~\ref{fig:a-fan}, we first generate a sequence of adversarial perturbations with diverse strengths to augment the intermediate features. Then, we inject perturbed feature statistics into clean features by feature normalization. In the end, the augmented and normalized features $f_{\mathrm{mix}}^{(i)}$ together with clean features $f_{\mathrm{clean}}$ are both utilized in the network training. In this way, adversarial training can be formulated as an effective regularization to improve the generalization ability of visual recognition models. The full algorithm is summarized in Algorithm~\ref{alg:a-fan}.

\vspace{-3mm}
\begin{algorithm}[H]
\caption{Adversarial Feature Augmentation and Normalization (A-FAN).}
\label{alg:a-fan}
\renewcommand{\algorithmicensure}{\textbf{Initialize:}}
\begin{algorithmic}[1]
\Ensure{$f(\mathbf x,\boldsymbol{\theta})$ is the visual recognition model, where $\boldsymbol{\theta}=(\boldsymbol{\theta}_1,\boldsymbol{\theta}_2)$. $f(\mathbf x,\boldsymbol{\theta}_1)$ are intermediate features.}
\State \textcolor{gray}{\# Generate adversarial augmented features}
\State Uniformly sample $k$ different perturbation strength $\{\epsilon^{(1)},\cdots,\epsilon^{(k)}\}$ from $[0,\mathcal{E}]$.
\State Generate adversarial perturbations $\delta_1(\epsilon^{(1)})$ with PGD, according to Equation~\ref{eq:attack_on_x} and~\ref{eq:afa}.
\State Apply $\delta_1(\epsilon^{(1)})$ to the intermediate features and obtain adversarial features $f_{\mathrm{adv}}^{(1)}$.
\For {$\epsilon^{(i)}\in\{\epsilon^{(2)},\cdots,\epsilon^{(k)}\}$}
\State Generate other augmented features $f_{\mathrm{adv}}^{(2)}, \cdots, f_{\mathrm{adv}}^{(k)}$ via the efficient implementation in Section~\ref{sec:afa}.
\EndFor
\State \textcolor{gray}{\# Generate adversarial normalized features}
\State Calculate the feature statistics $\mu_{\mathrm{clean}}$, $\sigma_{\mathrm{clean}}$ and $\{\mu_{\mathrm{adv}}^{(i)},\sigma_{\mathrm{adv}}^{(i)}\}_{i=1}^{k}$ with batch normalization \cite{ioffe2015batch}.
\For {$i\in\{1,2,\cdots,k\}$}
\State Inject adversarial feature statistics $\mu_{\mathrm{adv}}^{(i)},\sigma_{\mathrm{adv}}^{(i)}$ into clean features $f_{\mathrm{clean}}$ via the normalization, and obtain normalized features $f_{\mathrm{mix}}^{(i)}$, according to Equation~\ref{eq:afn}.
\EndFor
\State Feed normalized features to the model and compute the complete objective of A-FAN in Equation~\ref{eq:finalloss}.
\\ \Return{Training objective $\mathcal{L}_{\mathrm{A-FAN}}$}
\end{algorithmic}
\end{algorithm}

After incorporating adversarial feature augmentation and normalization, the complete training objective of A-FAN can be computed as follows:
\begin{align}\label{eq:finalloss}
    %\begin{array}{l}
        %\displaystyle 
        f_{\mathrm{A-FAN}} \Def \mathcal{L}_{\mathrm{clean}} + \lambda\textstyle{\sum_{i=1}^{k}}\mathcal{L}_{\mathrm{adv}}^{(i)}\,,
     %\end{array}
\end{align}
where $\lambda=1$ is tuned by grid search. 

\section{A-FAN on Image Classification}
\paragraph{Datasets and Backbones.} We consider three representative datasets for image classification: CIFAR-10, CIFAR-100 \cite{krizhevsky2009learning}, and ImageNet \cite{deng2009imagenet}. In our experiments, the original training datasets are randomly split into $90\%$ training and $10\%$ validation. The early stopping technique is applied to find the top-performing checkpoints on the validation set. Then, the selected checkpoints are evaluated on the test set to report the performance. The hyperparameters are tuned by grid search, which are quite stable from validation to test sets based on our observations, including PGD steps, step size $\alpha$, the layers to introduce adversarial perturbations, and the number of perturbations with different strength levels. We evaluate large backbone networks (ResNet-18/50/101/152 \cite{he2016deep}, EfficientNet-B0 \cite{tan2019efficientnet}) on ImageNet, and test smaller backbones (ResNet-20s/56s) as well on CIFAR-10 and CIFAR-100. More details about training and evaluation are provided in Section~\ref{apx:class_details}. 

\begin{table} [t!]
\caption{Standard testing accuracy (SA\%) of ResNet-20s/56s on CIFAR-10 and CIFAR-100. Baseline denotes the standard training without A-FAN. \textcolor{red}{$\uparrow$} indicates the improvement over SA compared to the corresponding baseline in standard training.}
% \vspace{1mm}
\label{tab:cifar}
\centering
\resizebox{0.47\textwidth}{!}{
\begin{tabular}{c|cc|cc}
\toprule
\multirow{2}{*}{Settings} & \multicolumn{2}{c}{CIFAR-10} & \multicolumn{2}{c}{CIFAR-100} \\ \cmidrule(lr){2-3} \cmidrule(lr){4-5}
& Baseline & A-FAN & Baseline & A-FAN \\ \midrule
ResNet-20s & 91.25 & 92.52 (\textcolor{red}{$\uparrow$ 1.27}) & 66.92 & 67.89 (\textcolor{red}{$\uparrow$ 0.97}) \\
ResNet-56s & 93.59 & 94.82 (\textcolor{red}{$\uparrow$ 1.23}) & 71.22 & 72.36 (\textcolor{red}{$\uparrow$ 1.14}) \\
\bottomrule
\end{tabular}}
\vspace{-1mm}
\end{table}

\begin{table} [t!]
\caption{Standard testing accuracy (SA\%) of ResNet-18/50/101/152 and EfficientNet-B0 on the ImageNet dataset.}
% \vspace{1mm}
\label{tab:imagenet}
\centering
\resizebox{0.38\textwidth}{!}{
\begin{tabular}{c|cc}
\toprule
\multirow{2}{*}{Settings} & \multicolumn{2}{c}{ImageNet}\\ \cmidrule(lr){2-3}
 & Baseline & Baseline + A-FAN  \\ \midrule
ResNet-18 & 69.38 & 70.25 (\textcolor{red}{$\uparrow$ 0.87})   \\
ResNet-50 & 75.21 & 76.33 (\textcolor{red}{$\uparrow$ 1.12}) \\
ResNet-101 & 77.10 & 78.14 (\textcolor{red}{$\uparrow$ 1.04}) \\
ResNet-152 & 78.31 & 78.69 (\textcolor{red}{$\uparrow$ 0.38}) \\
EfficientNet-B0 & 77.04 & 77.50 (\textcolor{red}{$\uparrow$ 0.46})\\
\bottomrule
\end{tabular}}
\vspace{-3mm}
\end{table}

\vspace{-3mm}
\paragraph{CIFAR and ImageNet Results.} We apply PGD-5 and PGD-1 to augment the feature embeddings in the last block with adversarial perturbations for CIFAR and ImageNet models, respectively. A series of adversarial augmented features are crafted with three different strengths uniformly sampled from [0,$\alpha$], where the step size $\alpha=0.5/255$. Table~\ref{tab:cifar} and Table~\ref{tab:imagenet} present the standard testing accuracy of diverse models on CIFAR-10, CIFAR-100 and ImageNet. Comparing the standard training (i.e., Baseline) with our proposed A-FAN, here are the main observations: 
\begin{itemize} 
    \item A-FAN obtains a consistent and substantial improvement over standard accuracy, \textit{e.g.}, $1.27\%$ on CIFAR-10 with ResNet-20s, $1.14\%$ on CIFAR-100 with ResNet-56s, $1.12\%$ and $0.46\%$ on ImageNet with ResNet-50 and EfficientNet-B0. This suggests that training with augmented and normalized features generated by A-FAN effectively enhances the generalization of deep networks. We hypothesize that it is because adversarial perturbed features are treated as an implicit regularization, leading to better solutions for network training.
    \item Shallow ResNets benefit more from A-FAN than deep ResNets (\textit{e.g.}, $1.12\%$ on ResNet-50 vs. $0.38\%$ on ResNet-152). A possible reason is that the performance of standard trained deep ResNets is already saturated, leaving little room for improvement. 
\end{itemize}
Furthermore, we notice that A-FAN advocates different steps of PGD to achieve superior performance on diverse datasets. More ablation analyses can be found in Section~\ref{sec:ablation}. Meanwhile, although the robust testing accuracy is not the focus of A-FAN, we report it for completeness in Section~\ref{apx:class_details}.

\begin{table} [t!]
\caption{Running time per epoch and standard testing accuracy (SA\%) comparison across Baseline, AdvProp, and A-FAN.}
\label{tab:time}
\centering
\resizebox{0.47\textwidth}{!}{
\begin{tabular}{c|cc|cc}
\toprule
\multirow{2}{*}{Settings} & \multicolumn{2}{c}{ResNet-18 on CIFAR-10} & \multicolumn{2}{|c}{EfficientNet-B0 on ImageNet} \\
\cmidrule{2-3} \cmidrule{4-5}
& SA & Time & SA & Time \\ \midrule 
Baseline & 94.30 & 23s & 77.00 & 2628s\\
AdvProp & 94.52 (\textcolor{red}{$\uparrow$ 0.22}) & 123s & 77.60 (\textcolor{red}{$\uparrow$ 0.60}) & 13352s\\
A-FAN & 94.67 (\textcolor{red}{$\uparrow$ 0.37}) & 56s & 77.50 (\textcolor{red}{$\uparrow$ 0.50}) & 6237s\\
\bottomrule
\end{tabular}}
\vspace{-5mm}
\end{table}

\vspace{-3mm}
\paragraph{A-FAN vs. AdvProp.} We compare A-FAN with AdvProp \cite{xie2020adversarial} on CIFAR-10 with ResNet-18, and on ImageNet with EfficientNet-B0 \cite{tan2019efficientnet}, as presented in Table~\ref{tab:time}. CIFAR-10 models are trained on a single GTX1080 Ti GPU. ImageNet (batch size 256) experiments are conducted on $2\times$ Quadro RTX 6000 GPUs with 24G$\times$2 memory in total. Since for generating feature-level perturbations, only a \textit{partial} backpropagation to the target intermediate layer is needed which brings computational saving. The results also confirm our intuition that proposed A-FAN as an effective and efficient alternative for pixel-level adversarial augmentations (e.g., AdvProp), achieves competitive performance with much more less computational cost (i.e., less running time). 

\section{A-FAN on Object Detection}
\paragraph{Datasets and Backbones.} We evaluate A-FAN on Pascal VOC2007 \cite{everingham2010pascal} and COCO2017 \cite{lin2014microsoft} for object detection. COCO2017 is a large-scale dataset with more than ten times of data than Pascal VOC2007. In our experiments, we choose the widely-used framework, Faster RCNN \cite{ren2015faster}, for detection tasks. It is worth mentioning that the proposed A-FAN approach can be directly plugged into other detection frameworks without any change, which is left to future work. We conduct experiments with both ResNet-50 \cite{he2016deep} and ResNet-101 \cite{he2016deep} as backbone networks. More details about training and evaluation are be found in Section~\ref{apx:detection_details}. 

\vspace{-2mm}
\paragraph{Pascal VOC and COCO Results.} Results are presented in Table~\ref{table:sota_detection}. All hyperparameters of A-FAN are tuned by grid search, including PGD steps, step size $\alpha$, the layers to introduce adversarial feature augmentations, and the number of perturbations with different strength levels. We find that utilizing PGD-1 to generate adversarial feature perturbations in the last layer of backbone and ROI networks of Faster RCNN, achieves the most promising performance. We adopt $\alpha=0.3/255$ for Pascal VOC2007 and $\alpha=0.5/255$ for COCO2017. For both datasets, a series of adversarial augmented features are crafted with five different strengths uniformly sampled from [0,$\alpha$]. To evaluate the robustness (\emph{i.e.}, robust AP) of detection model \cite{li2018robust,xie2017adversarial}, PGD-10 attack with $\alpha=0.3/255$ and $\epsilon=2.0/255$ is applied.

\begin{table}[t!]
\centering
\caption{Performance of object detection on Pascal VOC2007 and COCO2017 datasets. Faster RCNN is equipped with ResNet-50/ResNet-101 backbone networks, respectively. Robustness is evaluated on the adversarial perturbed images \cite{li2018robust,xie2017adversarial} via PGD-10.}
\label{table:sota_detection}
\begin{adjustbox}{width=0.47\textwidth}
\begin{tabular}{l|cc|cc}
\toprule
\multirow{2}{*}{COCO2017} & \multicolumn{2}{c|}{ResNet-50} & \multicolumn{2}{c}{ResNet-101}\\ \cmidrule{2-3} \cmidrule{4-5}
& Baseline & A-FAN  & Baseline & A-FAN  \\ \midrule
AP (\%) & 33.20 & 33.85 & 36.21 & 37.05\\
AP$_{50}$ (\%) & 53.92 & 54.73 & 56.90 & 57.31\\
AP$_{75}$ (\%) & 35.83 & 36.54 & 39.40 & 40.22\\
Robust AP (\%) & 0.00 & 0.50 & 0.20 & 0.66\\
\bottomrule
\end{tabular}
\end{adjustbox}
\begin{adjustbox}{width=0.47\textwidth}
\begin{tabular}{l|cc|cc}
\toprule
\multirow{2}{*}{Pascal VOC2007} & \multicolumn{2}{c}{ResNet-50} & \multicolumn{2}{|c}{ResNet-101}\\ \cmidrule{2-3} \cmidrule{4-5}
 & Baseline & A-FAN & Baseline & A-FAN   \\ \midrule
mAP (\%) & 73.96 & 75.38 & 74.32 & 75.71 \\
Robust mAP (\%) & 0.86 & 2.43 & 1.71 & 3.85 \\
\bottomrule
\end{tabular}
\end{adjustbox}
\vspace{-3mm}
\end{table}

Table~\ref{table:sota_detection} summarizes the results of the baseline (\emph{i.e.}, standard training) and A-FAN. More results with different training settings are provided in Section~\ref{apx:detection_res}. Comparing standard training with our proposed A-FAN mechanism, several major observations can be drawn:
\begin{itemize}
    \item A-FAN consistently achieves substantial performance improvement across multiple backbones on diverse datasets. Specifically, A-FAN gains $0.65\%$/$0.84\%$ AP with ResNet-50/ResNet-101 on COCO2007, and $1.42\%$/$1.39\%$ mAP with ResNet-50/ResNet-101 on Pascal VOC2007. This demonstrates that training with adversarially augmented and normalized features crafted via A-FAN significantly boosts the generalization of detection models. A possible reason is that utilizing adversarially perturbed features as an implicit regularization for training leads to better generalization. 
    \item Detectors trained on small-scale dataset benefits more from A-FAN. For example, Faster RCNN with ResNet-50 backbone obtains an almost double mAP\footnote{AP$_{50}$ shares the same meaning as mAP in VOC datasets \cite{9102805}} boost (\emph{i.e.}, $1.42\%$ vs. $0.81\%$) on VOC2007 than on COCO2017. It comes as no surprise that adversarially augmented and normalized features can be regarded as data augmentation in the embedding space and therefore perform more effectively on small-scale datasets \cite{shorten2019survey}. We also notice that Faster RCNN with both shallow and deep ResNets gets a similar degree of improvement.
    \item Besides the enhanced generalization ability, detectors trained with A-FAN also receive better robustness, improved by $0.46\%\sim 0.50\%$ robust AP on COCO2017 and $1.57\%\sim 2.14\%$ robust mAP on Pascal VOC2007. Although the improved robustness still cannot hold a candle to adversarially trained models \cite{dai2016r,ren2016faster,lin2017feature}, it is an extra bonus from A-FAN.
    \item A-FAN can achieve similar improvements, compared to other previous/identical data augmentations (e.g., \cite{zoph2019learning}).
\end{itemize}

\section{A-FAN on Semantic Segmentation}
\paragraph{Datasets and Backbones.} We validate the effectiveness of A-FAN on Pascal VOC2007 \cite{everingham2010pascal}, Pascal VOC2012 \cite{everingham2015pascal}, and Cityspaces \cite{cordts2016cityscapes} for semantic segmentation. Among these commonly used datasets, Cityspaces is a large-scale datasets with more than ten times of data than Pascal VOC2007/2012. In our experiments, the popular framework DeepLab V3+ \cite{chen2018encoder} with ResNet-50 \cite{he2016deep} and ResNet-101 \cite{he2016deep} as backbone networks, is adopted for segmentation tasks. Note that A-FAN can also be directly plugged into other segmentation frameworks without any change, which is left to future work. More details are referred to Section~\ref{apx:segmentation_details}. 

\begin{table}[t!]
\centering
\caption{Performance of object detection on Pascal VOC2007 and COCO2017 datasets. Faster RCNN is equipped with ResNet-50/ResNet-101 backbone networks. Robustness is evaluated on adversarially perturbed images \cite{shen2019advspade} via PGD-10.}
\label{table:sota_segmentation}
\begin{adjustbox}{width=0.47\textwidth}
\begin{tabular}{l|cc|cc}
\toprule
\multirow{2}{*}{Pascal VOC2012 } & \multicolumn{2}{c|}{ResNet-50} & \multicolumn{2}{c}{ResNet-101}\\ \cmidrule{2-3} \cmidrule{4-5}
& Baseline & A-FAN  & Baseline & A-FAN  \\ \midrule
mIOU (\%) & 71.20 & 72.21 & 73.65 & 74.91 \\
Robust mIOU (\%) & 10.84 & 12.07 & 9.75 & 11.01 \\
\bottomrule
\end{tabular}
\end{adjustbox}
\begin{adjustbox}{width=0.47\textwidth}
\begin{tabular}{l|cc|cc}
\toprule
\multirow{2}{*}{ResNet-50} & \multicolumn{2}{c}{Pascal VOC2007} & \multicolumn{2}{|c}{Cityspaces}\\ \cmidrule{2-3} \cmidrule{4-5}
& Baseline & A-FAN  & Baseline & A-FAN  \\ \midrule
mIOU (\%) & 61.51 & 62.83 & 76.00 & 76.43 \\
Robust mIOU (\%) & 6.77 & 7.06 & 0.51 & 1.11 \\
\bottomrule
\end{tabular}
\end{adjustbox}
\vspace{-3mm}
\end{table}

\vspace{-3mm}
\paragraph{Pascal VOC and Cityspaces Results.} Results are collected in Table~\ref{table:sota_segmentation}. We adopt PGD-1 to craft adversarially augmented features with three different perturbation strengths (sampled from [0,$\alpha$]) in the last layer of backbone and the decoder networks of DeepLab V3+ with $\alpha=1.0/255,0.4/255,0.3/255$ for Pascal VOC2007, Pascal VOC2012 and Cityspaces, respectively. All hyperparameters are tuned by grid search. PGD-10 with $\alpha=1.0/255$ and $\epsilon=8.0/255$ is employed to measure robustness (\emph{i.e.}, Robust mIOU) of segmentation models \cite{shen2019advspade}.

From the results in Table~\ref{table:sota_segmentation}, we observe that Deeplab V3+ gains substantial performance improvement from A-FAN, which is consistent with the observations on detection models. First, A-FAN enhances the generalization of segmentation models by $1.01\%$/$1.26\%$ mIOU with ResNet-50/ResNet-101 on Pascal VOC2012, $1.32\%$ with ResNet-50 on Pascal VOC2007, and $0.43\%$ mIOU with ResNet-50 on Cityspaces. Second, A-FAN improves Deeplab V3+ more on Pascal VOC2007/2012 than on Cityspaces (\emph{i.e.}, $1.01\%\sim1.32\%$ vs. $0.43\%$), where the former two datasets only have one-tenth amount of data compared to Cityspaces. Third, Training with A-FAN yields moderate robustness improvement (\emph{i.e.}, $0.29\%\sim1.25\%$ robust mIOU) for segmentation models.

\section{Ablation Study and Analyses} \label{sec:ablation}
Due to the limited space, more ablation results and analyses can be found in Section~\ref{apx:class_res},~\ref{apx:detection_res} and~\ref{apx:class_res} for classification, detection and segmentation models, respectively.

\vspace{-4mm}
\paragraph{Augmentation vs. Normalization} To verify the effects of adversarial feature augmentation (AFA) and adversarial feature normalization (AFN) in A-FAN, we incrementally evaluate each module on CIFAR-10 for image classification, Pascal VOC2007 for object detection, and Pascal VOC2012 for semantic segmentation. As shown in Table~\ref{table:ablation_cls_more} and Table~\ref{table:ablation}, AFA improves the baseline by $0.86\%$ SA/$1.13\%$ AP/$0.89\%$ mIOU for classification, detection and segmentation, respectively. The combination of the two modules, AFA and AFN, gains a further performance boost by $0.37\%$ SA/$0.29\%$ AP/$0.12\%$ mIOU on CIFAR-10, Pascal VOC2007 and VOC2012. These results demonstrate that each proposed component contributes to improving the generalization ability of detection and segmentation models, and AFA plays a dominant role in ameliorating performance. 

\begin{table}[t!]
\centering
\caption{Ablation study of A-FAN on Pascal VOC2007 and Pascal VOC2012 for detection and segmentation, respectively. AFA: adversarial feature augmentation; AFN: adversarial feature normalization (\emph{i.e.}, A-FAN = AFA + AFN). ResNet-50 backbone is used here. \textcolor{red}{$\uparrow$} indicates performance improvement compared to the corresponding baseline. Classification results are in Table~\ref{table:ablation_cls_more}.}
\label{table:ablation}
\begin{adjustbox}{width=0.47\textwidth}
\begin{tabular}{l|c|c}
\toprule
\multirow{2}{*}{Settings} & \multicolumn{1}{c}{Detection} & \multicolumn{1}{|c}{Segmentation}\\ \cmidrule{2-3}
 & AP (\%) & mIOU (\%) \\ \midrule
Baseline & 73.96 & 71.20 \\ 
\qquad \qquad + AFA & 75.09 (\textcolor{red}{$\uparrow$ 1.13}) & 72.09 (\textcolor{red}{$\uparrow$ 0.89}) \\
 \qquad \qquad + AFA + AFN & 75.38 (\textcolor{red}{$\uparrow$ 1.42}) & 72.21 (\textcolor{red}{$\uparrow$ 1.01})\\ \midrule
A-FAN on Backbone & 75.06 (\textcolor{red}{$\uparrow$ 1.10}) & 71.98 (\textcolor{red}{$\uparrow$ 0.78}) \\
A-FAN on ROI/Decoder & 74.68 (\textcolor{red}{$\uparrow$ 0.72}) & 71.71  (\textcolor{red}{$\uparrow$ 0.51})  \\ 
A-FAN on Both & 75.38 (\textcolor{red}{$\uparrow$ 1.42}) & 72.21 (\textcolor{red}{$\uparrow$ 1.01}) \\
\bottomrule
\end{tabular}
\end{adjustbox}
\vspace{-5mm}
\end{table}

\begin{figure}[t!] 
\centering
\includegraphics[width=0.8\linewidth]{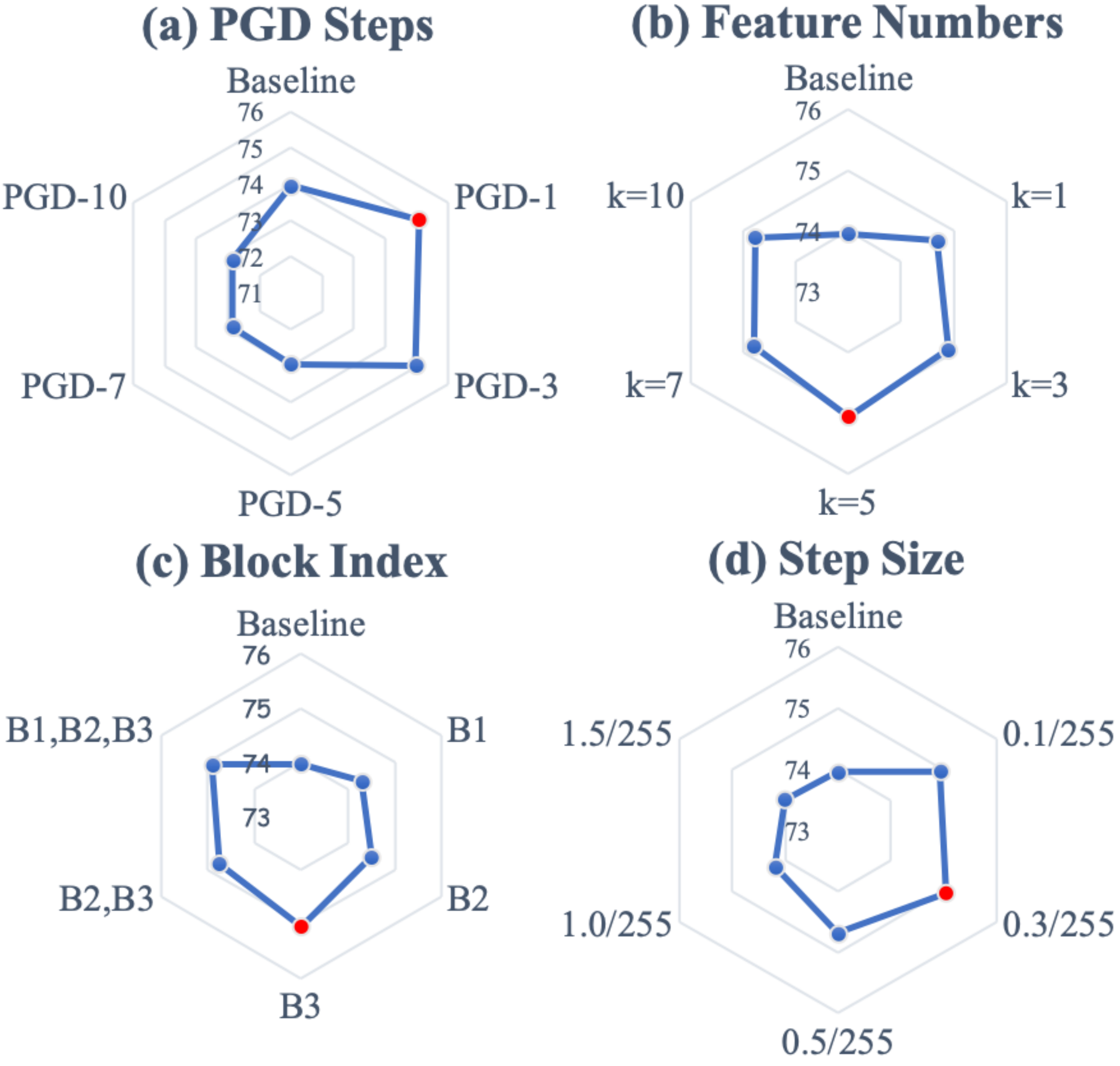}
\vspace{-1mm}
\caption{Ablation study on the location and strength of introducing A-FAN to detection models. Results are on Pascal VOC2007 dataset. (a) PGD steps used in the generation of adversarial perturbations; (b) The number of augmented features, ($k$ in Equation~\ref{eq:afa_app}); (c) The location to apply A-FAN, \emph{e.g.}, B1 means that A-FAN is applied to features from the first residual blocks in the ResNet backbone; (d) Step size $\alpha$ that controls the strength of crafted perturbations. The \textcolor{red}{red} points represent settings with top performance.}
\label{fig:aba_detect}
\vspace{-3mm}
\end{figure}

\vspace{-4mm}
\paragraph{Effects on Backbone v.s. ROI/Decoder.} In general, detection and segmentation models can be divided into backbone and task-specific modules (\emph{e.g.}, RPN/ROI in Faster RCNN \cite{ren2016faster} and ASPP/Decoder in Deeplab V3+). Our proposed A-FAN can be introduced to either or both modules as shown in detailed ablations in Table~\ref{table:ablation}. We observe that applying A-FAN to backbone networks (\textcolor{red}{$\uparrow$}$1.10\%$ AP/\textcolor{red}{$\uparrow$}$0.78\%$ mIOU) gains more generalization improvement than ROI/Decoder modules (\textcolor{red}{$\uparrow$}$0.72\%$ AP/\textcolor{red}{$\uparrow$}$0.51\%$ mIOU) for detection and segmentation. Incorporating A-FAN on both backbone and task-specific modules always enjoys extra performance boost, compared to applying either one alone.

\vspace{-3mm}
\paragraph{Effects of Location and Strength.} The performance gain from A-FAN is determined by the location and strength of generated adversarial perturbations.  Figure~\ref{fig:aba_cls},~\ref{fig:aba_detect} and~\ref{fig:aba_seg} illustrate a comprehensive control study to investigate these relevant factors. Without losing generality, these ablation experiments and analyses are performed on backbone networks. When studying one of the factors, we choose the best configuration for the other factors.

To identify the proper location for A-FAN operation we inject feature perturbations to different blocks (\emph{e.g.}, B1) or some combination of blocks (\emph{e.g.}, B2,B3), as presented in Figure~\ref{fig:aba_cls} (c) for classification,~\ref{fig:aba_detect} (c) for detection, and~\ref{fig:aba_seg} (c) for segmentation. We notice that applying A-FAN to features from the last block (\emph{i.e.}, B3 or B4) obtains the best performance, while introducing A-FAN to multiple blocks degrades generalization.

The strength of A-FAN includes the number of PGD steps and the step size $\alpha$ for generating adversarial features, and the number of augmented features with different perturbation strengths ($k$ in Equation~\ref{eq:afa_app}), as shown in Figure~\ref{fig:aba_cls} (a),(b),(c) for classification,~\ref{fig:aba_detect} (a),(b),(c) for detection, and~\ref{fig:aba_seg} (a),(b),(c) for segmentation. Experiments show that \{ResNet-18, Faster RCNN, Deeplab V3+\} gains more from A-FAN with \{PGD-5,PGD-1,PGD-1\}, step size $\alpha=\{\frac{1.0}{255},\frac{0.3}{255},\frac{0.4}{255}\}$, and \{3,5,3\} augmented features with different perturbation strength. These systematic evaluations reveal that: $(i)$ weak (\emph{e.g.}, $\alpha=0.1/255$) adversarial perturbed features contribute marginal generalization improvements; $(ii)$ excessively strong (\emph{e.g.}, PGD-10, $\alpha=4.0/255$) A-FAN incurs performance deterioration. In summary, we observe that a proper configuration for A-FAN usually produces high-quality augmented and normalized features, realizing enhanced visual recognition models.

% \begin{figure}[!ht] 
% \centering
% \includegraphics[width=1\linewidth]{Figs/Ablation_detect.pdf}
% \caption{Ablation study on the location and strength of introducing A-FAN to detection models. Results are on Pascal VOC2007 dataset. (a) PGD steps used in the generation of adversarial perturbations; (b) The number of augmented features, ($k$ in Equation~\ref{eq:afa_app}); (c) The location to apply A-FAN, \emph{e.g.}, B1 means that A-FAN is applied to features from the first residual blocks in the ResNet backbone; (d) Step size $\alpha$ that controls the strength of crafted perturbations. The \textcolor{red}{red} points represent settings with top performance.}
% \label{fig:aba_detect}
% \end{figure}

\vspace{-3mm}
\paragraph{Comparing A-FAN with Random Noise.} One straight-froward approach to augment feature embeddings is injecting random noise. Here we replace the generated adversarial noise in our proposed mechanism with a randomly sampled noise from Gaussian distribution $\mathcal{N}(0,\alpha^2)$. As shown in Table~\ref{table:noise}, AFA+AFN (\emph{i.e.}, A-FAN) achieves a larger performance gain than Random Noise+AFN, suggesting that gradient-based crafted feature augmentation benefits more to the generalization ability of visual recognition models.

\begin{table}[t!]
\centering
\caption{Performance comparison between adversarial feature perturbations with the strength $\alpha$ and random noise sampled from a Gaussian distribution $\mathcal{N}(0,\alpha^2)$. Results are reported on CIFAR-10 (with ResNet-56s), Pascal VOC2007, and Pascal VOC2012 for classification, detection, and segmentation, respectively. \textcolor{red}{$\uparrow$}/\textcolor{blue}{$\downarrow$} indicates performance improvement/degradation compared to baseline.}
\label{table:noise}
\begin{adjustbox}{width=0.47\textwidth}
\begin{tabular}{l|c|c|c}
\toprule
\multirow{2}{*}{Settings} & \multicolumn{1}{c}{CIFAR-10} & \multicolumn{1}{c}{VOC2007} & \multicolumn{1}{|c}{VOC2012}\\ \cmidrule{2-4}
 & SA (\%) & AP (\%) & mIOU (\%) \\ \midrule
Random Noise + AFN & 93.36 (\textcolor{blue}{$\downarrow$ 0.23}) & 73.91 (\textcolor{blue}{$\downarrow$ 0.05}) & 71.23  (\textcolor{red}{$\uparrow$ 0.03}) \\ 
AFA + AFN (i.e. A-FAN) & 94.82 (\textcolor{red}{$\uparrow$ 1.23})  & 75.38 (\textcolor{red}{$\uparrow$ 1.42}) & 72.21 (\textcolor{red}{$\uparrow$ 1.01})\\ 
\bottomrule
\end{tabular}
\end{adjustbox}
\vspace{-1mm}
\end{table}

\vspace{-1.5mm}
\paragraph{Visualization.} Figure~\ref{fig:vis_cls},~\ref{fig:vis} and~\ref{fig:vis_seg} provide visualization of adversarially augmented features and normalized features generated by A-FAN. Features are collected via applying A-FAN to classification, detection and segmentation models on ImageNet, Pascal VOC2007 and VOC2012 datasets, respectively. Visualization of classification models can be found in Section~\ref{apx:class_res}. For better visualization, we use features from the first block of backbone networks and further enlarge the magnitude of adversarial perturbations by $\times20$ times. We notice that normalizing features by injecting adversarial statistics into clean features, seems to neutralize the excessively generated adversarial noise. It offers an explanation for the extra performance improvement by adversarial feature normalization.

\section{Conclusion and Discussion}
In this paper, we present A-FAN, an enhanced adversarial training method to improve image classification, object detection, and semantic segmentation. By generating a series of adversarial perturbations with different strengths on feature embeddings, and fusing adversarial feature statistics with clean features, A-FAN substantially boost the generalization ability of various models across multiple representative datasets, such as CIFAR-10/100, ImageNet, Pascal VOC2007/2012, COCO2017 and Cityspaces. 
%Comprehensive ablation studies validate the effectiveness of each proposed component. 
For future work, we would like to extend A-FAN to more tasks and provide theoretical understanding of A-FAN.

\clearpage

{\small
\bibliographystyle{ieee_fullname}
\bibliography{ALFA_CV}

\begin{thebibliography}{10}\itemsep=-1pt

\bibitem{arnab2018robustness}
Anurag Arnab, Ondrej Miksik, and Philip~HS Torr.
\newblock On the robustness of semantic segmentation models to adversarial
  attacks.
\newblock In {\em CVPR}, pages 888--897, 2018.

\bibitem{ba2016layer}
Jimmy~Lei Ba, Jamie~Ryan Kiros, and Geoffrey~E Hinton.
\newblock Layer normalization.
\newblock {\em arXiv preprint arXiv:1607.06450}, 2016.

\bibitem{bartlett2002rademacher}
Peter~L Bartlett and Shahar Mendelson.
\newblock Rademacher and gaussian complexities: Risk bounds and structural
  results.
\newblock {\em Journal of Machine Learning Research}, 3(Nov):463--482, 2002.

\bibitem{carlini2017towards}
Nicholas Carlini and David Wagner.
\newblock Towards evaluating the robustness of neural networks.
\newblock In {\em 2017 ieee symposium on security and privacy (sp)}, pages
  39--57. IEEE, 2017.

\bibitem{chen2018encoder}
Liang-Chieh Chen, Yukun Zhu, George Papandreou, Florian Schroff, and Hartwig
  Adam.
\newblock Encoder-decoder with atrous separable convolution for semantic image
  segmentation.
\newblock In {\em ECCV}, pages 801--818, 2018.

\bibitem{chen2020adversarial}
Tianlong Chen, Sijia Liu, Shiyu Chang, Yu Cheng, Lisa Amini, and Zhangyang
  Wang.
\newblock Adversarial robustness: From self-supervised pre-training to
  fine-tuning.
\newblock In {\em Proceedings of the IEEE/CVF Conference on Computer Vision and
  Pattern Recognition}, pages 699--708, 2020.

\bibitem{chen2021robust}
Tianlong Chen, Zhenyu Zhang, Sijia Liu, Shiyu Chang, and Zhangyang Wang.
\newblock Robust overfitting may be mitigated by properly learned smoothening.
\newblock In {\em International Conference on Learning Representations}, 2021.

\bibitem{chen2019multi}
Zitian Chen, Yanwei Fu, Yinda Zhang, Yu-Gang Jiang, Xiangyang Xue, and Leonid
  Sigal.
\newblock Multi-level semantic feature augmentation for one-shot learning.
\newblock {\em IEEE Transactions on Image Processing}, 28(9):4594--4605, 2019.

\bibitem{cohen2019certified}
Jeremy~M Cohen, Elan Rosenfeld, and J~Zico Kolter.
\newblock Certified adversarial robustness via randomized smoothing.
\newblock {\em arXiv preprint arXiv:1902.02918}, 2019.

\bibitem{cordts2016cityscapes}
Marius Cordts, Mohamed Omran, Sebastian Ramos, Timo Rehfeld, Markus Enzweiler,
  Rodrigo Benenson, Uwe Franke, Stefan Roth, and Bernt Schiele.
\newblock The cityscapes dataset for semantic urban scene understanding.
\newblock In {\em CVPR}, pages 3213--3223, 2016.

\bibitem{croce2020reliable}
Francesco Croce and Matthias Hein.
\newblock Reliable evaluation of adversarial robustness with an ensemble of
  diverse parameter-free attacks.
\newblock In {\em ICML}, 2020.

\bibitem{cubuk2018autoaugment}
Ekin~D Cubuk, Barret Zoph, Dandelion Mane, Vijay Vasudevan, and Quoc~V Le.
\newblock Autoaugment: Learning augmentation policies from data.
\newblock {\em arXiv preprint arXiv:1805.09501}, 2018.

\bibitem{dai2016r}
Jifeng Dai, Yi Li, Kaiming He, and Jian Sun.
\newblock R-fcn: Object detection via region-based fully convolutional
  networks.
\newblock In {\em NeurIPS}, pages 379--387, 2016.

\bibitem{deng2009imagenet}
Jia Deng, Wei Dong, Richard Socher, Li-Jia Li, Kai Li, and Li Fei-Fei.
\newblock Imagenet: A large-scale hierarchical image database.
\newblock In {\em CVPR}, pages 248--255, 2009.

\bibitem{dhillon2018stochastic}
Guneet~S Dhillon, Kamyar Azizzadenesheli, Zachary~C Lipton, Jeremy Bernstein,
  Jean Kossaifi, Aran Khanna, and Anima Anandkumar.
\newblock Stochastic activation pruning for robust adversarial defense.
\newblock {\em arXiv preprint arXiv:1803.01442}, 2018.

\bibitem{dziugaite2016study}
Gintare~Karolina Dziugaite, Zoubin Ghahramani, and Daniel~M Roy.
\newblock A study of the effect of jpg compression on adversarial images.
\newblock {\em arXiv preprint arXiv:1608.00853}, 2016.

\bibitem{everingham2015pascal}
Mark Everingham, SM~Ali Eslami, Luc Van~Gool, Christopher~KI Williams, John
  Winn, and Andrew Zisserman.
\newblock The pascal visual object classes challenge: A retrospective.
\newblock {\em International journal of computer vision}, 111(1):98--136, 2015.

\bibitem{everingham2010pascal}
Mark Everingham, Luc Van~Gool, Christopher~KI Williams, John Winn, and Andrew
  Zisserman.
\newblock The pascal visual object classes (voc) challenge.
\newblock {\em International journal of computer vision}, 88(2):303--338, 2010.

\bibitem{gan2020large}
Zhe Gan, Yen-Chun Chen, Linjie Li, Chen Zhu, Yu Cheng, and Jingjing Liu.
\newblock Large-scale adversarial training for vision-and-language
  representation learning.
\newblock In {\em NeurIPS}, 2020.

\bibitem{Detectron2018}
Ross Girshick, Ilija Radosavovic, Georgia Gkioxari, Piotr Doll\'{a}r, and
  Kaiming He.
\newblock Detectron.
\newblock \url{https://github.com/facebookresearch/detectron}, 2018.

\bibitem{goodfellow2014explaining}
Ian~J Goodfellow, Jonathon Shlens, and Christian Szegedy.
\newblock Explaining and harnessing adversarial examples.
\newblock {\em arXiv preprint arXiv:1412.6572}, 2014.

\bibitem{guo2017countering}
Chuan Guo, Mayank Rana, Moustapha Cisse, and Laurens Van Der~Maaten.
\newblock Countering adversarial images using input transformations.
\newblock {\em arXiv preprint arXiv:1711.00117}, 2017.

\bibitem{he2016deep}
Kaiming He, Xiangyu Zhang, Shaoqing Ren, and Jian Sun.
\newblock Deep residual learning for image recognition.
\newblock In {\em CVPR}, pages 770--778, 2016.

\bibitem{hendrycks2019augmix}
Dan Hendrycks, Norman Mu, Ekin~Dogus Cubuk, Barret Zoph, Justin Gilmer, and
  Balaji Lakshminarayanan.
\newblock Augmix: A simple data processing method to improve robustness and
  uncertainty.
\newblock In {\em International Conference on Learning Representations}, 2019.

\bibitem{hofmann2008kernel}
Thomas Hofmann, Bernhard Sch{\"o}lkopf, and Alexander~J Smola.
\newblock Kernel methods in machine learning.
\newblock {\em The annals of statistics}, pages 1171--1220, 2008.

\bibitem{hu2019triple}
Ting-Kuei Hu, Tianlong Chen, Haotao Wang, and Zhangyang Wang.
\newblock Triple wins: Boosting accuracy, robustness and efficiency together by
  enabling input-adaptive inference.
\newblock In {\em International Conference on Learning Representations}, 2019.

\bibitem{ioffe2015batch}
Sergey Ioffe and Christian Szegedy.
\newblock Batch normalization: Accelerating deep network training by reducing
  internal covariate shift.
\newblock {\em arXiv preprint arXiv:1502.03167}, 2015.

\bibitem{8852250}
M. {Ishii} and A. {Sato}.
\newblock Training deep neural networks with adversarially augmented features
  for small-scale training datasets.
\newblock In {\em 2019 International Joint Conference on Neural Networks
  (IJCNN)}, pages 1--8, 2019.

\bibitem{jiang2020robust}
Ziyu Jiang, Tianlong Chen, Ting Chen, and Zhangyang Wang.
\newblock Robust pre-training by adversarial contrastive learning.
\newblock {\em arXiv preprint arXiv:2010.13337}, 2020.

\bibitem{kakade2008complexity}
Sham~M Kakade, Karthik Sridharan, and Ambuj Tewari.
\newblock On the complexity of linear prediction: Risk bounds, margin bounds,
  and regularization.
\newblock 2008.

\bibitem{koltchinskii2002empirical}
Vladimir Koltchinskii, Dmitry Panchenko, et~al.
\newblock Empirical margin distributions and bounding the generalization error
  of combined classifiers.
\newblock {\em Annals of statistics}, 30(1):1--50, 2002.

\bibitem{krizhevsky2009learning}
Alex Krizhevsky et~al.
\newblock Learning multiple layers of features from tiny images.
\newblock 2009.

\bibitem{kurakin2016adversarial}
Alexey Kurakin, Ian Goodfellow, and Samy Bengio.
\newblock Adversarial examples in the physical world.
\newblock {\em arXiv preprint arXiv:1607.02533}, 2016.

\bibitem{li2020feature}
Boyi Li, Felix Wu, Ser-Nam Lim, Serge Belongie, and Kilian~Q Weinberger.
\newblock On feature normalization and data augmentation.
\newblock {\em arXiv preprint arXiv:2002.11102}, 2020.

\bibitem{li2019positional}
Boyi Li, Felix Wu, Kilian~Q Weinberger, and Serge Belongie.
\newblock Positional normalization.
\newblock In {\em Advances in Neural Information Processing Systems}, pages
  1622--1634, 2019.

\bibitem{li1998sphering}
Guoying Li and Jian Zhang.
\newblock Sphering and its properties.
\newblock {\em Sankhy{\=a}: The Indian Journal of Statistics, Series A}, pages
  119--133, 1998.

\bibitem{visualloss}
Hao Li, Zheng Xu, Gavin Taylor, Christoph Studer, and Tom Goldstein.
\newblock Visualizing the loss landscape of neural nets.
\newblock In {\em Neural Information Processing Systems}, 2018.

\bibitem{li2018exploring}
Yuezun Li, Xiao Bian, Ming-Ching Chang, and Siwei Lyu.
\newblock Exploring the vulnerability of single shot module in object detectors
  via imperceptible background patches.
\newblock {\em arXiv preprint arXiv:1809.05966}, 2018.

\bibitem{li2018robust}
Yuezun Li, Daniel Tian, Ming-Ching Chang, Xiao Bian, and Siwei Lyu.
\newblock Robust adversarial perturbation on deep proposal-based models.
\newblock {\em arXiv preprint arXiv:1809.05962}, 2018.

\bibitem{liao2018defense}
Fangzhou Liao, Ming Liang, Yinpeng Dong, Tianyu Pang, Xiaolin Hu, and Jun Zhu.
\newblock Defense against adversarial attacks using high-level representation
  guided denoiser.
\newblock In {\em CVPR}, pages 1778--1787, 2018.

\bibitem{lin2017feature}
Tsung-Yi Lin, Piotr Doll{\'a}r, Ross Girshick, Kaiming He, Bharath Hariharan,
  and Serge Belongie.
\newblock Feature pyramid networks for object detection.
\newblock In {\em CVPR}, pages 2117--2125, 2017.

\bibitem{lin2014microsoft}
Tsung-Yi Lin, Michael Maire, Serge Belongie, James Hays, Pietro Perona, Deva
  Ramanan, Piotr Doll{\'a}r, and C~Lawrence Zitnick.
\newblock Microsoft coco: Common objects in context.
\newblock In {\em ECCV}, pages 740--755. Springer, 2014.

\bibitem{liu2016ssd}
Wei Liu, Dragomir Anguelov, Dumitru Erhan, Christian Szegedy, Scott Reed,
  Cheng-Yang Fu, and Alexander~C Berg.
\newblock Ssd: Single shot multibox detector.
\newblock In {\em ECCV}, pages 21--37. Springer, 2016.

\bibitem{liu2018towards}
Xuanqing Liu, Minhao Cheng, Huan Zhang, and Cho-Jui Hsieh.
\newblock Towards robust neural networks via random self-ensemble.
\newblock In {\em ECCV}, pages 369--385, 2018.

\bibitem{liu2018adv}
Xuanqing Liu, Yao Li, Chongruo Wu, and Cho-Jui Hsieh.
\newblock Adv-bnn: Improved adversarial defense through robust bayesian neural
  network.
\newblock {\em arXiv preprint arXiv:1810.01279}, 2018.

\bibitem{liu2018dpatch}
Xin Liu, Huanrui Yang, Ziwei Liu, Linghao Song, Hai Li, and Yiran Chen.
\newblock Dpatch: An adversarial patch attack on object detectors.
\newblock {\em arXiv preprint arXiv:1806.02299}, 2018.

\bibitem{lu2017adversarial}
Jiajun Lu, Hussein Sibai, and Evan Fabry.
\newblock Adversarial examples that fool detectors.
\newblock {\em arXiv preprint arXiv:1712.02494}, 2017.

\bibitem{madry2017towards}
Aleksander Madry, Aleksandar Makelov, Ludwig Schmidt, Dimitris Tsipras, and
  Adrian Vladu.
\newblock Towards deep learning models resistant to adversarial attacks.
\newblock {\em arXiv preprint arXiv:1706.06083}, 2017.

\bibitem{montavon2012neural}
Gr{\'e}goire Montavon, G Orr, and Klaus-Robert M{\"u}ller.
\newblock Neural networks-tricks of the trade second edition.
\newblock 2012.

\bibitem{nakkiran2019adversarial}
Preetum Nakkiran.
\newblock Adversarial robustness may be at odds with simplicity.
\newblock {\em arXiv preprint arXiv:1901.00532}, 2019.

\bibitem{raghunathan2018semidefinite}
Aditi Raghunathan, Jacob Steinhardt, and Percy~S Liang.
\newblock Semidefinite relaxations for certifying robustness to adversarial
  examples.
\newblock In {\em NeurIPS}, pages 10877--10887, 2018.

\bibitem{raghunathan2019adversarial}
Aditi Raghunathan, Sang~Michael Xie, Fanny Yang, John~C Duchi, and Percy Liang.
\newblock Adversarial training can hurt generalization.
\newblock In {\em ICMLW}, 2019.

\bibitem{ren2015faster}
Shaoqing Ren, Kaiming He, Ross Girshick, and Jian Sun.
\newblock Faster r-cnn: Towards real-time object detection with region proposal
  networks.
\newblock In {\em NeurIPS}, pages 91--99, 2015.

\bibitem{ren2016faster}
Shaoqing Ren, Kaiming He, Ross Girshick, and Jian Sun.
\newblock Faster r-cnn: Towards real-time object detection with region proposal
  networks.
\newblock {\em IEEE transactions on pattern analysis and machine intelligence},
  39(6):1137--1149, 2016.

\bibitem{schmidt2018adversarially}
Ludwig Schmidt, Shibani Santurkar, Dimitris Tsipras, Kunal Talwar, and
  Aleksander Madry.
\newblock Adversarially robust generalization requires more data.
\newblock In {\em NeurIPS}, pages 5014--5026, 2018.

\bibitem{scholkopf1996incorporating}
Bernhard Sch{\"o}lkopf, Chris Burges, and Vladimir Vapnik.
\newblock Incorporating invariances in support vector learning machines.
\newblock In {\em ICANN}, pages 47--52. Springer, 1996.

\bibitem{shen2019advspade}
Guangyu Shen, Chengzhi Mao, Junfeng Yang, and Baishakhi Ray.
\newblock Advspade: Realistic unrestricted attacks for semantic segmentation.
\newblock {\em arXiv preprint arXiv:1910.02354}, 2019.

\bibitem{shorten2019survey}
Connor Shorten and Taghi~M Khoshgoftaar.
\newblock A survey on image data augmentation for deep learning.
\newblock {\em Journal of Big Data}, 6(1):60, 2019.

\bibitem{simard1993efficient}
Patrice Simard, Yann LeCun, and John~S Denker.
\newblock Efficient pattern recognition using a new transformation distance.
\newblock In {\em NeurIPS}, pages 50--58, 1993.

\bibitem{stutz2019disentangling}
David Stutz, Matthias Hein, and Bernt Schiele.
\newblock Disentangling adversarial robustness and generalization.
\newblock In {\em CVPR}, pages 6976--6987, 2019.

\bibitem{sun2019towards}
Ke Sun, Zhanxing Zhu, and Zhouchen Lin.
\newblock Towards understanding adversarial examples systematically:
  {Exploring} data size, task and model factors.
\newblock {\em arXiv preprint arXiv:1902.11019}, 2019.

\bibitem{szegedy2013intriguing}
Christian Szegedy, Wojciech Zaremba, Ilya Sutskever, Joan Bruna, Dumitru Erhan,
  Ian Goodfellow, and Rob Fergus.
\newblock Intriguing properties of neural networks.
\newblock {\em arXiv preprint arXiv:1312.6199}, 2013.

\bibitem{tan2019efficientnet}
Mingxing Tan and Quoc~V Le.
\newblock Efficientnet: Rethinking model scaling for convolutional neural
  networks.
\newblock {\em arXiv preprint arXiv:1905.11946}, 2019.

\bibitem{tsipras2018robustness}
Dimitris Tsipras, Shibani Santurkar, Logan Engstrom, Alexander Turner, and
  Aleksander Madry.
\newblock Robustness may be at odds with accuracy.
\newblock In {\em ICLR}, 2019.

\bibitem{ulyanov2016instance}
Dmitry Ulyanov, Andrea Vedaldi, and Victor Lempitsky.
\newblock Instance normalization: The missing ingredient for fast stylization.
\newblock {\em arXiv preprint arXiv:1607.08022}, 2016.

\bibitem{8578674}
R. {Volpi}, P. {Morerio}, S. {Savarese}, and V. {Murino}.
\newblock Adversarial feature augmentation for unsupervised domain adaptation.
\newblock In {\em CVPR}, pages 5495--5504, 2018.

\bibitem{wang2019improving}
Dilin Wang, Chengyue Gong, and Qiang Liu.
\newblock Improving neural language modeling via adversarial training.
\newblock {\em arXiv preprint arXiv:1906.03805}, 2019.

\bibitem{wei2019improved}
Colin Wei and Tengyu Ma.
\newblock Improved sample complexities for deep networks and robust
  classification via an all-layer margin.
\newblock {\em arXiv preprint arXiv:1910.04284}, 2019.

\bibitem{wei2018transferable}
Xingxing Wei, Siyuan Liang, Ning Chen, and Xiaochun Cao.
\newblock Transferable adversarial attacks for image and video object
  detection.
\newblock {\em arXiv preprint arXiv:1811.12641}, 2018.

\bibitem{wu2018group}
Yuxin Wu and Kaiming He.
\newblock Group normalization.
\newblock In {\em Proceedings of the European conference on computer vision
  (ECCV)}, pages 3--19, 2018.

\bibitem{xie2020adversarial}
Cihang Xie, Mingxing Tan, Boqing Gong, Jiang Wang, Alan~L Yuille, and Quoc~V
  Le.
\newblock Adversarial examples improve image recognition.
\newblock In {\em CVPR}, pages 819--828, 2020.

\bibitem{xie2017adversarial}
Cihang Xie, Jianyu Wang, Zhishuai Zhang, Yuyin Zhou, Lingxi Xie, and Alan
  Yuille.
\newblock Adversarial examples for semantic segmentation and object detection.
\newblock In {\em ICCV}, pages 1369--1378, 2017.

\bibitem{xu2017feature}
Weilin Xu, David Evans, and Yanjun Qi.
\newblock Feature squeezing: Detecting adversarial examples in deep neural
  networks.
\newblock {\em arXiv preprint arXiv:1704.01155}, 2017.

\bibitem{yun2019cutmix}
Sangdoo Yun, Dongyoon Han, Seong~Joon Oh, Sanghyuk Chun, Junsuk Choe, and
  Youngjoon Yoo.
\newblock Cutmix: Regularization strategy to train strong classifiers with
  localizable features.
\newblock In {\em ICCV}, pages 6023--6032, 2019.

\bibitem{zhang2017mixup}
Hongyi Zhang, Moustapha Cisse, Yann~N Dauphin, and David Lopez-Paz.
\newblock mixup: Beyond empirical risk minimization.
\newblock {\em arXiv preprint arXiv:1710.09412}, 2017.

\bibitem{zhang2019theoretically}
Hongyang Zhang, Yaodong Yu, Jiantao Jiao, Eric~P Xing, Laurent~El Ghaoui, and
  Michael~I Jordan.
\newblock Theoretically principled trade-off between robustness and accuracy.
\newblock {\em arXiv preprint arXiv:1901.08573}, 2019.

\bibitem{zhang2020contextual}
Hantao Zhang, Wengang Zhou, and Houqiang Li.
\newblock Contextual adversarial attacks for object detection.
\newblock In {\em ICME}, pages 1--6. IEEE, 2020.

\bibitem{9102805}
H. {Zhang}, W. {Zhou}, and H. {Li}.
\newblock Contextual adversarial attacks for object detection.
\newblock In {\em 2020 IEEE International Conference on Multimedia and Expo
  (ICME)}, pages 1--6, 2020.

\bibitem{zhang2019feature}
Yinghui Zhang, Bo Sun, Yongkang Xiao, Rong Xiao, and YunGang Wei.
\newblock Feature augmentation for imbalanced classification with conditional
  mixture wgans.
\newblock {\em Signal Processing: Image Communication}, 75:89--99, 2019.

\bibitem{zhu2019freelb}
Chen Zhu, Yu Cheng, Zhe Gan, Siqi Sun, Tom Goldstein, and Jingjing Liu.
\newblock Freelb: Enhanced adversarial training for natural language
  understanding.
\newblock In {\em ICLR}, 2020.

\bibitem{zoph2019learning}
Barret Zoph, Ekin~D Cubuk, Golnaz Ghiasi, Tsung-Yi Lin, Jonathon Shlens, and
  Quoc~V Le.
\newblock Learning data augmentation strategies for object detection.
\newblock {\em arXiv preprint arXiv:1906.11172}, 2019.

\end{thebibliography}
}

\clearpage

\appendix
\renewcommand{\thepage}{S\arabic{page}}  
\renewcommand{\thesection}{S\arabic{section}}   
\renewcommand{\thetable}{S\arabic{table}}   
\renewcommand{\thefigure}{S\arabic{figure}}

\section{More Methodology Details}
\subsection{Efficient Implementation of A-FAN} 
In practice, we offer an efficient implementation to compute Equation~\ref{eq:afa_app} for A-FAN with PGD-1. Since we by default \cite{madry2017towards,xie2020adversarial} use $\epsilon=\frac{8.0}{255}$, for most A-FAN experiments with a small $\alpha$ and PGD-1, the PGD will degrade to simple gradient descent. Similar routines are also adopted in \cite{gan2020large}. In this sense, step size $\alpha$ is the only indicator for perturbation strength, satisfying $\boldsymbol{\delta}_i=\frac{\alpha^{(i)}}{\alpha^{(j)}}\boldsymbol{\delta}_j$ as demonstrated in Equation~\ref{eq:attack_on_x}. Thus, $f^{(i)}_{\mathrm{adv}}$ can be efficiently calculated by $f^{(i)}_{\mathrm{adv}}=f^{(1)}_{\mathrm{adv}}+(\frac{\alpha^{(i)}}{\alpha^{(1)}}-1)\boldsymbol{\delta}_1$, which merely requires to apply PGD once rather than $k$ times in Equation~\ref{eq:afa_app}, and then derives other augmented features with negligible extra cost. Note that it is not available for the multi-step PGD or large step sizes (e.g., $\alpha>\epsilon$).

\begin{table*}[t]
\centering
\caption{Details of training and evaluation. We use the standard implementations and hyperparameters in \cite{ren2015faster,chen2018encoder}. The evaluation metrics are also follow standards in \cite{ren2015faster,chen2018encoder}. Linear learning rate warm-up for $100$ iterations is applied.}
\label{table:settings}
\begin{adjustbox}{width=1\textwidth}
\begin{threeparttable}
\begin{tabular}{c|cc|ccc}
\toprule
\multirow{2}{*}{Datasets} & \multicolumn{2}{c}{Detection} & \multicolumn{3}{|c}{Segmentation} \\ \cmidrule{2-3} \cmidrule{4-6}
& Pascal VOC2007 & COCO2017 & Pascal VOC2007 & Pascal VOC2012 & Cityspaces \\ \midrule
Batch Size & $8$ & $8$ & $4$ & $4$ & $4$ \\ \midrule
Iterations & $11,250$ & $180,000$ & $10,000$ & $30,000$ & $30,000$ \\ \midrule
Init. Learning Rate & 0.008 & 0.01 & 0.01 & 0.01 & 0.1\\ \midrule
Learning Rate Decay &  $\times0.1$ at $6,250$, $8,750$ & $\times0.1$ at $120,000$, $160,000$ & \multicolumn{2}{c}{Polynomial w. power 0.9} & Polynomial w. power 0.9 \\ \midrule
Optimizer & \multicolumn{2}{c|}{SGD with momentum 0.9 and weight decay $5\times10^{-4}$} & \multicolumn{3}{c}{SGD with momentum 0.9 and weight decay $1\times10^{-4}$}\\ \midrule
Eval. Metric & mAP & AP, AP$_{50}$, AP$_{75}$ & mIOU & mIOU & mIOU \\
\bottomrule
\end{tabular}
\end{threeparttable}
\end{adjustbox}
\end{table*}

\section{More Implementation Details}
\subsection{More A-FAN on Image Classification} \label{apx:class_details}
\paragraph{Training Details and Evaluation Metrics.} For network training on CIFAR-10 and CIFAR-100, we adopt an SGD optimizer with a momentum of $0.9$, weight decay of $5\times10^{-4}$, and batch size of $128$ for $200$ epochs. The learning rate starts from $0.1$ and decays to one-tenth at $50$-th and $150$-th epochs. We also perform a linear learning rate warm-up in the first $200$ iterations. For ImageNet experiments, following the official setting in Pytorch repository,\footnote{https://github.com/pytorch/examples/tree/master/imagenet} we train deep networks for $90$ epochs with a batch size of $512$, and the learning rate decay at $30$-th and $60$-th epoch. The SGD optimizer is adopted with a momentum of $0.9$ and a weight decay of $1\times10^{-4}$. We evaluate the generalization ability of a network with Standard Testing Accuracy (\textbf{SA}), which represents image classification accuracy on the original clean test dataset.

% \paragraph{Perturbation Generation.}  For generated adversarial feature embeddings, we set the maximum perturbation magnitude $\epsilon$ to be unbounded, since there are no explicit constraints for feature perturbations, and the effect of tuning $\epsilon$ can be absorbed by adjusting PGD steps and step size $\alpha$. 

% In practice, the magnitude of crafted feature perturbation steadily stays in a range from $0.97$ to $1.10$ under the $\ell_2$ norm. Adversarial perturbations usually are applied to the normalized feature from batch normalization.

\subsection{More A-FAN on Object Detection} \label{apx:detection_details}
\paragraph{Training and Evaluation Metrics.} For both detection, we use the datasets following \cite{ren2015faster}: in Pascal VOC2007, we use the train and validation sets for training, and evaluate on test set; In COCO2017, we train models on the train set and evaluate on the validation set. All other implementation details are provided in Table~\ref{table:settings}.

\subsection{More A-FAN on Semantic Segmentation} \label{apx:segmentation_details}
\paragraph{Training and Evaluation Metrics.} For both segmentation, we use the datasets following \cite{chen2018encoder}: in Pascal VOC2007, we use the train and validation sets for training, and evaluate on test set; In Pascal VOC2012 and Cityspaces, we train models on the train set and evaluate on the validation set. All other implementation details are provided in Table~\ref{table:settings}.

\section{More Experiments Results}

\subsection{More Classification Results} \label{apx:class_res}
\paragraph{Training with Full Training Sets.} For a sanity check, we also conduct experiments with ResNet-50 on the full ($100\%$) ImageNet training set, Baseline ($75.98\%$) vs. A-FAN ($76.63\%$). Equipped with A-FAN, it obtain $0.65\%$ improvements in terms of the standard accuracy. 

\paragraph{Augmentation vs. Normalization.} To verify the effects of adversarial feature augmentation (AFA) and adversarial feature normalization (AFN) in A-FAN, we incrementally evaluate each module on CIFAR-10 with ResNet-56s. As shown in Table~\ref{table:ablation_cls_more}, these results show that each proposed component contributes to improving the generalization ability of classification models, and AFA plays a dominant role in ameliorating performance. 

\begin{table}[!htb]
\centering
\caption{Ablation study of A-FAN on CIFAR-10 with ResNet-56s. AFA: adversarial feature augmentation; AFN: adversarial feature normalization (\emph{i.e.}, A-FAN = AFA + AFN). \textcolor{red}{$\uparrow$} indicates performance improvement compared to the baseline on corresponding dataset.}
\vspace{1mm}
\label{table:ablation_cls_more}
\begin{adjustbox}{width=0.40\textwidth}
\begin{tabular}{l|c}
\toprule
\multirow{2}{*}{Settings} & \multicolumn{1}{c}{Classification} \\ \cmidrule{2-2}
 & Standard Accuracy (\%) \\ \midrule
Baseline & 93.59 \\ 
\qquad \qquad + AFA & 94.45 (\textcolor{red}{$\uparrow$ 0.86}) \\
 \qquad \qquad + AFA + AFN & 94.82 (\textcolor{red}{$\uparrow$ 1.23}) \\
\bottomrule
\end{tabular}
\end{adjustbox}
\end{table}

\begin{figure}[!ht] 
\centering
\includegraphics[width=1\linewidth]{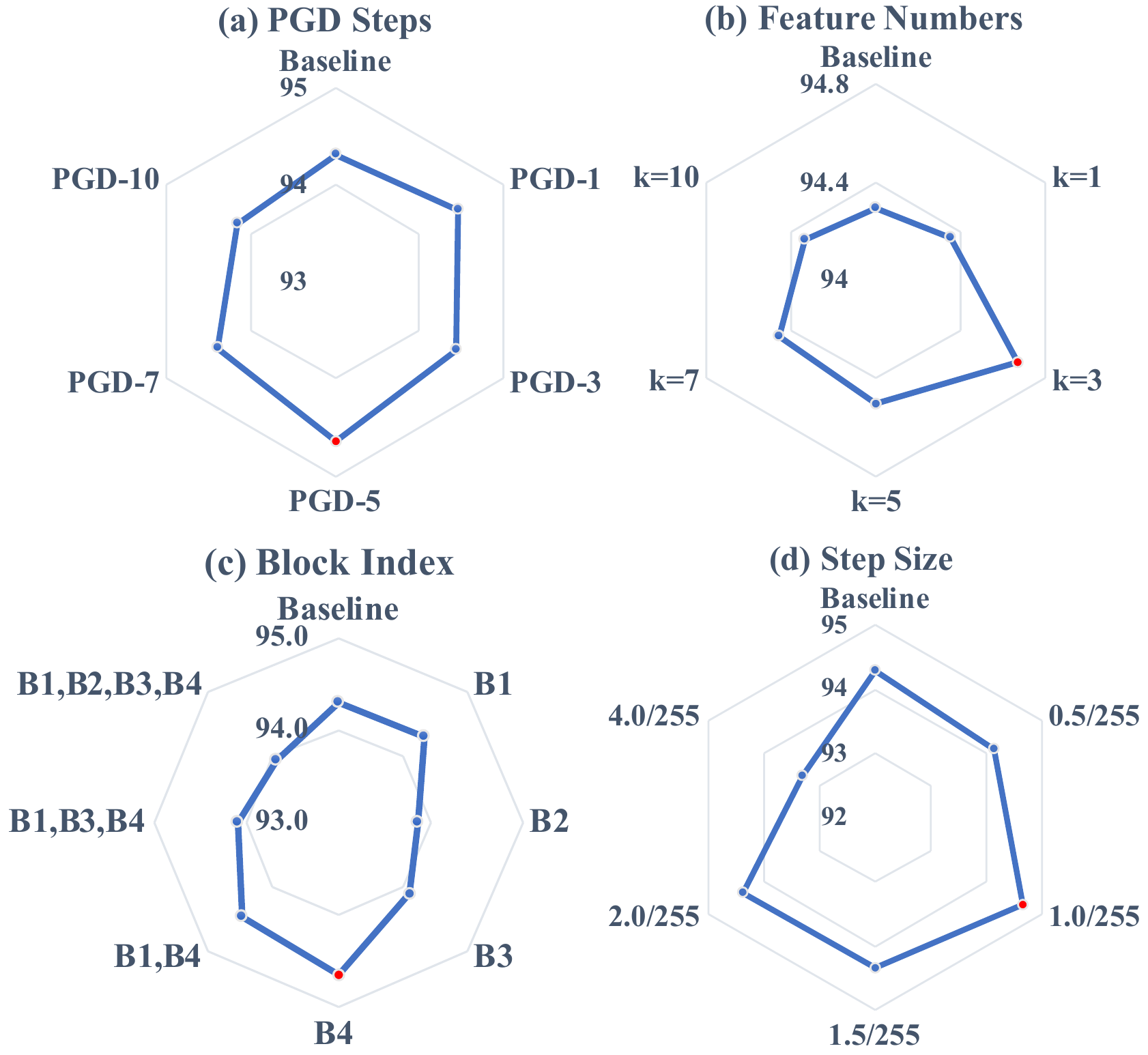}
\caption{Ablation study on the location and strength of introducing A-FAN to classification models. Results are on CIAFR-10 dataset with ResNet-18. (a) PGD steps used in the generation of adversarial perturbations; (b) The number of augmented features, ($k$ in Equation~\ref{eq:afa_app}); (c) The location to apply A-FAN, \emph{e.g.}, B1 means that A-FAN is applied to features from the first residual blocks in the ResNet backbone; (d) Step size $\alpha$ that controls the strength of crafted perturbations. The \textcolor{red}{red} points represent settings with top performance.}
\label{fig:aba_cls}
\end{figure}

\begin{figure}[t] 
\centering
\includegraphics[width=1\linewidth]{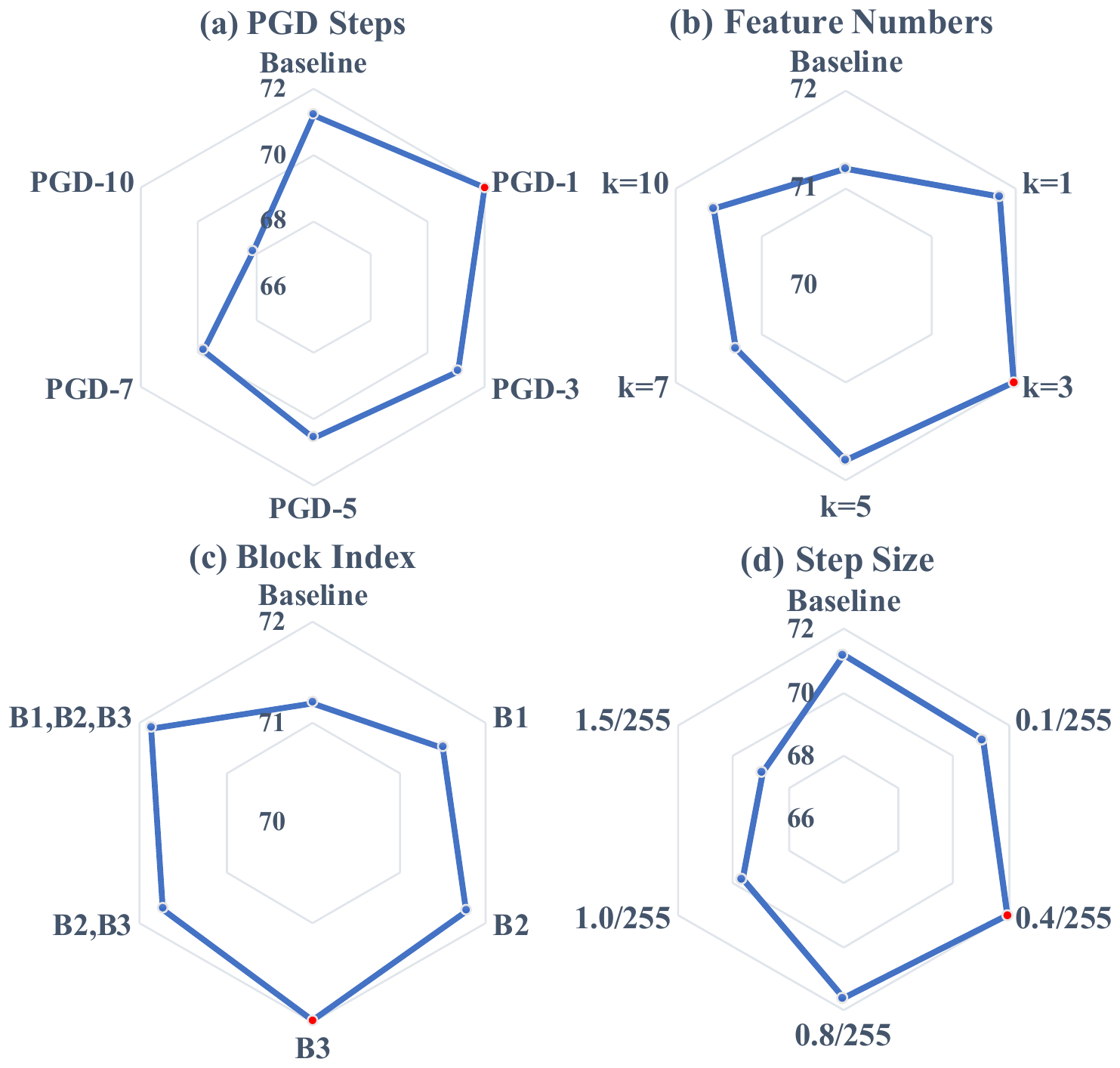}
\caption{The ablation study on the location and strength of introducing A-FAN to segmentation models. Results are on Pascal VOC2012 dataset. (a), (b), (c) and (d) share the same definitions as in Figure~\ref{fig:aba_detect}. The \textcolor{red}{red} points represent settings with top performance.}
\label{fig:aba_seg}
\end{figure}

\paragraph{Strength and Locations of A-FAN.} To understand the effect or the strength of injected adversarial perturbations, we implement ResNet-18 on CIFAR-10 and examine the performance across different step sizes, the number of PGD steps and augmented features. Figure~\ref{fig:aba_cls} shows that perturbing with PGD-1, step size $\alpha=1.0/255$ and augmenting three features with diverse perturbation strength achieve the superior performance, compared with other configurations. Generally, it shares similar observations to the ablation of detection (Figure~\ref{fig:aba_detect}) and segmentation (Figure~\ref{fig:aba_seg}) models. For the ablation of PGD steps, we implement ResNet-18 on ImageNet as well in Table~\ref{tab:pgdsteps}, which suggest A-FAN PGD-1 works the best for ImageNet.

Then, we analyze the effect of locations (\textit{i.e.}, where to apply A-FAN) via the typical backbone, ResNet-18 on CIFAR-10. In each setting, we present a detailed analysis on which layer and how many layers the feature embeddings should be adversarially augmented for achieving the best performance. Figure~\ref{fig:aba_cls} presents the layer preference of feature perturbations when applying A-FAN to different blocks or some combinations of blocks. We notice that introducing A-FAN to the last block achieves better standard accuracy, while the performance deteriorates after injecting A-FAN to multiple blocks.

\begin{table} [t!]
\caption{Standard testing accuracy (\%) on ImageNet datasets. As for ImageNet, we also perturb the last block features of ResNet-18 via PGD-1/3/5 and step size $\alpha=0.5/255$. The reference SA is $69.38\%$.}
\label{tab:pgdsteps}
\centering
\vspace{1mm}
\resizebox{0.47\textwidth}{!}{
\begin{tabular}{c|ccc}
\toprule
\multirow{2}{*}{Steps} & \multicolumn{3}{|c}{ImageNet} \\
\cmidrule{2-4}
& PGD-1 & PGD-3 & PGD-5 \\ \midrule 
A-FAN & 70.25 (\textcolor{red}{$\uparrow$ 0.87}) & 68.65 (\textcolor{blue}{$\downarrow$ 0.73}) & 67.42 (\textcolor{blue}{$\downarrow$ 1.96}) \\
\bottomrule
\end{tabular}}
\end{table}

\paragraph{Robust Performance of A-FAN.} Although the robust testing accuracy (RA) is not the focus of A-FAN, we report it for completeness. We implement the standard, A-FAN and the adversarial trained ResNet-18 networks on CIFAR-10. The adversarial trained model uses PGD-10 with step size $\alpha=\frac{2}{255}$ and $\epsilon=\frac{8}{255}$ for training. Then, PGD-20 with the same $\alpha$ and $\epsilon$ is applied to evaluate the robust performance of the three models. We observe that A-FAN trained models ($4.86\%$ RA) yield moderate robustness, compared to models from standard ($0.00\%$ RA) and adversarial ($50.72\%$ RA) trained models.

\begin{figure}[!ht] 
\centering
\includegraphics[width=1\linewidth]{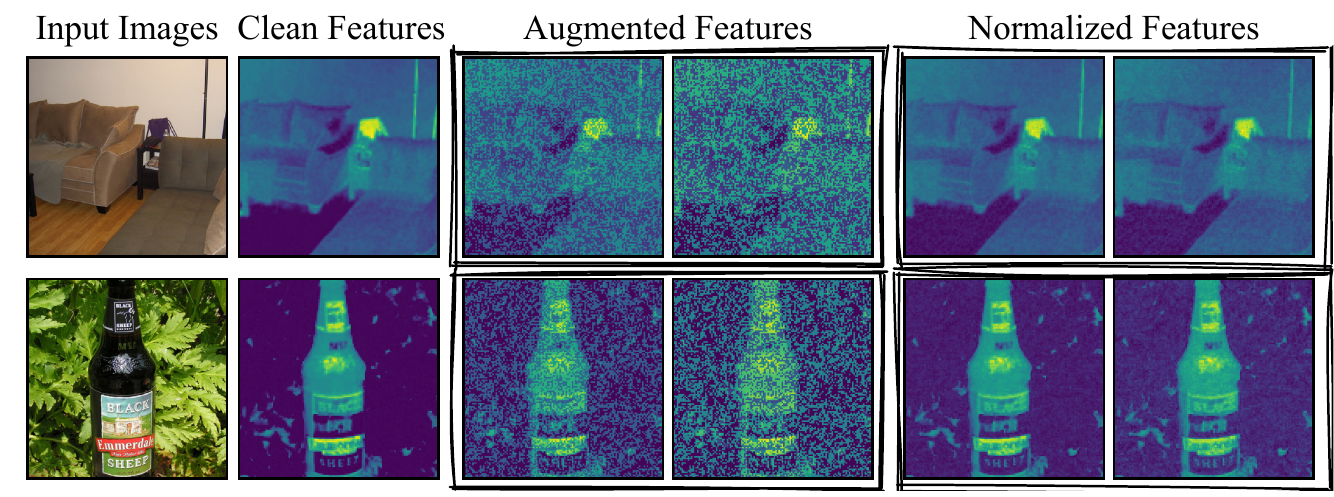}
\caption{Visualization of adversarially augmented and normalized features for classification models with A-FAN, using a trained ResNet-18. The fifth and sixth columns are normalized features of the third and forth columns, respectively.}
\label{fig:vis_cls}
\vspace{-4mm}
\end{figure}

\paragraph{Visualization.} Figure~\ref{fig:vis_cls} collects the visualization of adversarially augmented and normalized features for a trained ResNet-18 with A-FAN.

\section{More Objection Results} \label{apx:detection_res}
\paragraph{Addition Experiments of Detection on COCO2017.} Following another representative repository\footnote{\url{https://github.com/potterhsu/easy-faster-rcnn.pytorch}} for the Faster RCNN \cite{ren2015faster} implementation on COCO2017, we further verify the effectiveness of our proposed A-FAN. Table~\ref{table:settings_2} collects all detailed setup. As shown in Table~\ref{table:coco}, we observe that A-FAN boosts the baseline detection model by $0.96\%$ AP. Consistent performance gains in Table~\ref{table:coco} and the main text, reveals that A-FAN benefits detection models across diverse training configurations. 

\begin{table}[!ht]
\centering
\caption{Details of training and evaluation. We use the standard implementations and hyperparameters in the repository$^{\textcolor{red}{1}}$. The evaluation metrics are also follow standards in \cite{ren2015faster}. Linear learning rate warm-up for $100$ iterations is applied.}
\label{table:settings_2}
\begin{adjustbox}{width=0.47\textwidth}
\begin{threeparttable}
\begin{tabular}{c|c}
\toprule
\multirow{1}{*}{Datasets} & \multicolumn{1}{c}{Detection on COCO2017} \\ \midrule
Batch Size & $\boldsymbol{16}$  \\
Iterations  & $180,000$ \\
Init. Learning Rate  & 0.01 \\
Learning Rate Decay & $\times0.1$ at $120,000$, $160,000$\\
Optimizer & SGD with momentum 0.9 and weight decay $5\times10^{-4}$ \\
Eval. Metric & AP, AP$_{50}$, AP$_{75}$\\
\bottomrule
\end{tabular}
\end{threeparttable}
\end{adjustbox}
\end{table}

\begin{table}[!ht]
\centering
\caption{Performance of object detection on COCO2017 datasets. Faster RCNN is equipped with ResNet-101 backbone networks. Robustness are evaluated on the adversarial perturbed images \cite{li2018robust,xie2017adversarial} via PGD-10 with $\alpha=0.3/255$ and $\epsilon=2.0/255$.}
\label{table:coco}
\begin{adjustbox}{width=0.40\textwidth}
\begin{tabular}{l|cc}
\toprule
\multirow{2}{*}{Metrics} & \multicolumn{2}{c}{ResNet-101 on COCO2017} \\ \cmidrule{2-3} 
& Baseline & Baseline + A-FAN \\ \midrule
AP (\%) & 37.00 & 37.96 \\
AP$_{50}$ (\%) & 57.60 & 58.40  \\
AP$_{75}$ (\%) & 40.33 & 41.01  \\
Robust AP (\%) & 0.21 & 0.60  \\
\bottomrule
\end{tabular}
\end{adjustbox}
\end{table}

\paragraph{Comparison with Learned Data Augmentation (LDA) for Object Detection.} A recent work \cite{zoph2019learning} presents learned, specialized data augmentation policies to improve generalization performance for detection models. Although it is independent of our proposed feature-level adversarial augmentation, we still provide comparison experiments for a comprehensive investigation, as shown in Table~\ref{table:pascal}. Note that, for a fair comparison, we follow the exact same setting as \cite{zoph2019learning}. In addition to the detailed parameters, we combine the training sets of Pascal VOC 2007 and Pascal VOC 2012, and test the trained models on the Pascal VOC 2007 test set (4953 images). From Table~\ref{table:pascal}, we observe that both A-FAN and LDA obtain performance improvements by $3.66\%$ mAP and $2.70\%$ mAP, respectively. Achieved superior performance further validates the effectiveness of our proposed A-FAN.

\begin{table}[!ht]
\centering
\caption{Performance of object detection on Pascal VOC2007 datasets. Faster RCNN is equipped with ResNet-101 backbone networks. Robustness are evaluated on the adversarial perturbed images \cite{li2018robust,xie2017adversarial} via PGD-10 with $\alpha=0.3/255$ and $\epsilon=2.0/255$.}
\label{table:pascal}
\begin{adjustbox}{width=0.47\textwidth}
\begin{tabular}{l|ccc}
\toprule
\multirow{2}{*}{Metrics} & \multicolumn{3}{c}{ResNet-101 on Pascal VOC2007} \\ \cmidrule{2-4} 
& Baseline & Baseline + A-FAN & LDA\\ \midrule
mAP (\%) & 76.00 & 79.66 & 78.70\\
Robust mAP (\%) & 2.59 & 5.05 & - \\
\bottomrule
\end{tabular}
\end{adjustbox}
\vspace{-3mm}
\end{table}

\begin{figure}[!ht] 
\centering
\includegraphics[width=1\linewidth]{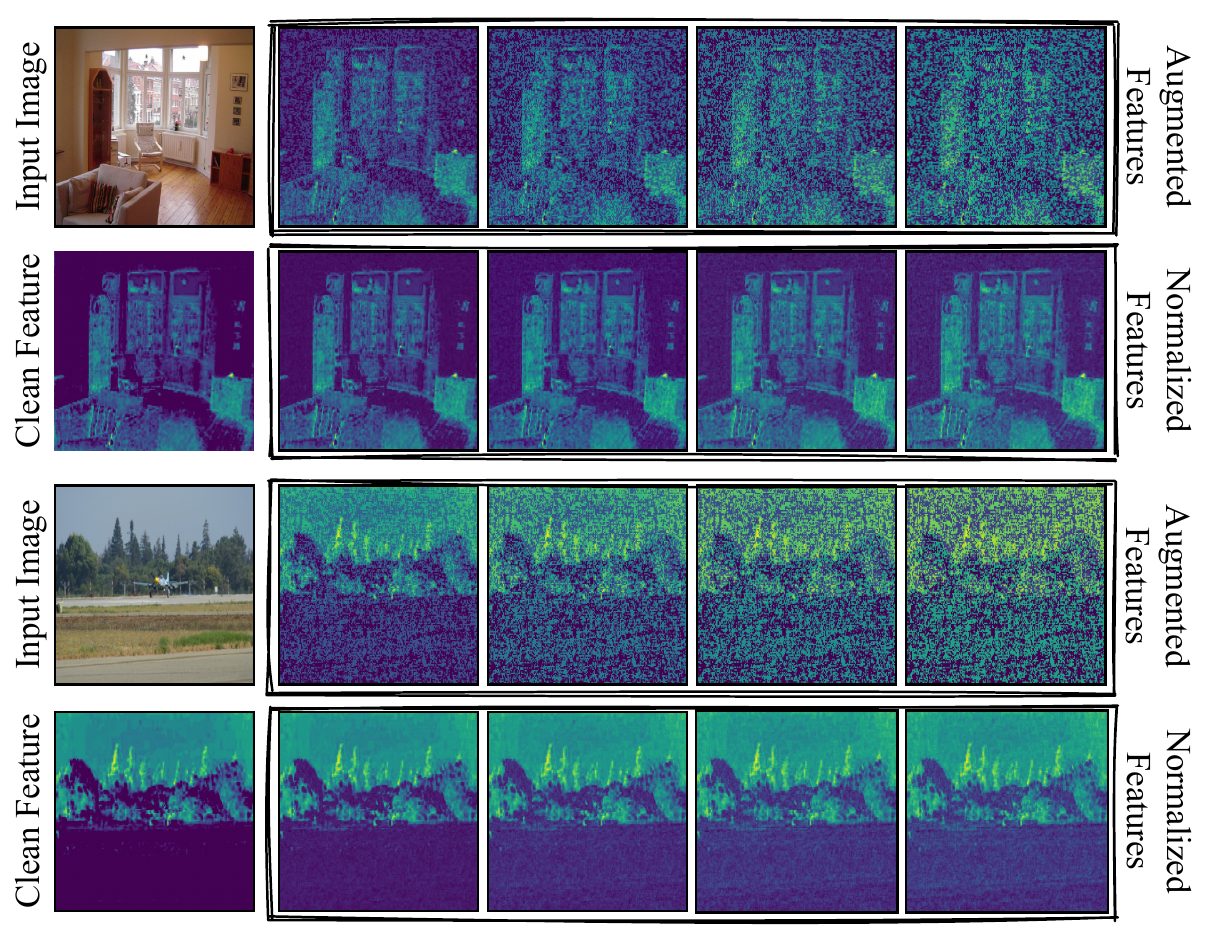}
\caption{Visualization of adversarial augmented and normalized features for detection models with A-FAN, using a trained Faster RCNN. The left column shows the input image and the corresponding clean feature. The remaining four columns, from left to right, present features with an increased perturbation strength; from up to bottom, it shows augmented and normalized features alternatively.}
\label{fig:vis}
% \vspace{-4mm}
\end{figure}

\paragraph{Visualization.} Figure~\ref{fig:vis} presents the visualization of adversarial augmented and normalized features for detection models with A-FAN, using a trained Faster RCNN.

\section{More Segmentation Results} \label{apex:seg_res}

\paragraph{Strength and Locations of A-FAN.} Figure~\ref{apex:seg_res} provides a comprehensive control study to investigate the relevant factors of A-FAN with segmentation models.

\begin{figure}[!ht] 
\centering
\includegraphics[width=1\linewidth]{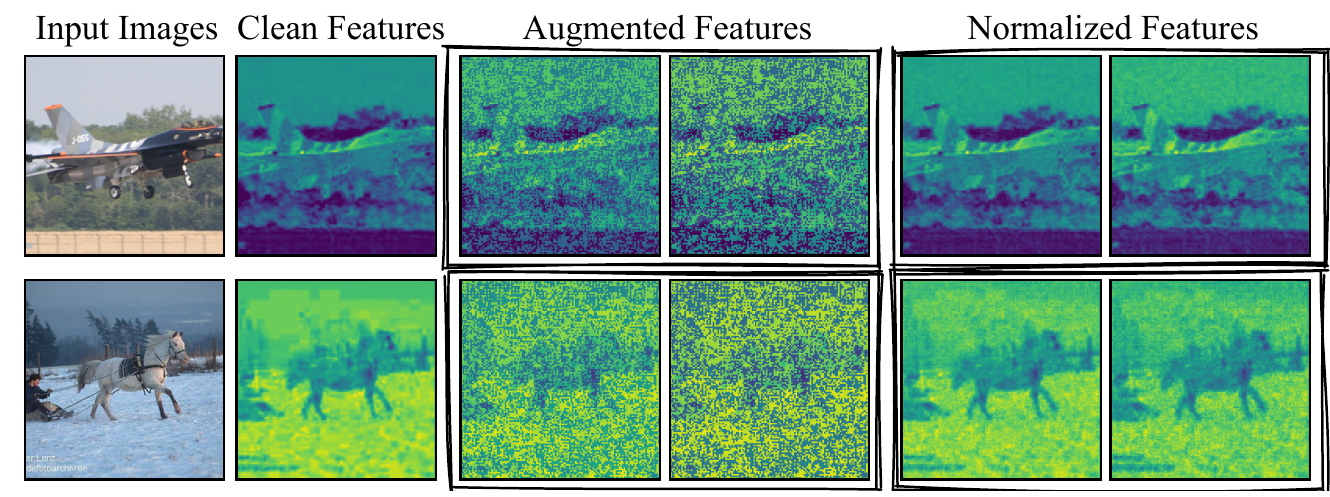}
\caption{Visualization of adversarially augmented and normalized features for segmentation models with A-FAN, using a trained Deeplab V3+. The fifth and sixth columns are normalized features of the third and forth columns, respectively.}
\label{fig:vis_seg}
\vspace{-4mm}
\end{figure}

\paragraph{Visualization.} Figure~\ref{fig:vis_seg} collects the visualization of adversarially augmented and normalized features for segmentation models with A-FAN, using a trained Deeplab V3+.

\end{document}